\PassOptionsToPackage{colorlinks=true,linkcolor=blue,citecolor=black,urlcolor=cyan,filecolor=black,pagebackref=false,hypertexnames=false}{hyperref}
\PassOptionsToPackage{sort&compress}{natbib}
\documentclass{article}

\usepackage[table]{xcolor}
\usepackage[final]{corl_2026}
\usepackage{amsmath}
\usepackage{amssymb}
\usepackage{cases}
\usepackage{graphicx}
\usepackage{float}
\usepackage{booktabs}

\usepackage{amsthm}

\theoremstyle{definition}

\newtheorem{problem}{Problem}
\AfterEndEnvironment{problem}{\noindent\ignorespaces}

\usepackage{mdframed} 

\usepackage{subcaption}
\usepackage{wrapfig}
\usepackage{placeins}
\usepackage{multirow}
\usepackage{mathtools}
\usepackage{empheq}
\usepackage{bbm}
\usepackage{xspace}

\usepackage[shortlabels]{enumitem}
\usepackage{tabularray}
\usepackage[final]{microtype} 
\usepackage{tabularx}

\usepackage{tikz} 
\usepackage{tkz-graph}
\usepackage{tkz-berge}
\usetikzlibrary{backgrounds,fit,shapes,snakes,arrows,shapes.geometric,positioning}
\usetikzlibrary{intersections,patterns,shapes.misc}
\usetikzlibrary{decorations.pathmorphing}
\usepackage{pgfplots}
\pgfplotsset{compat=1.18}

\tikzstyle{block} = [rectangle, rounded corners, minimum width=3cm, minimum height=1cm,text centered, draw=black, fill=red!30]
\tikzstyle{new} = [rectangle, rounded corners, minimum width=1cm, minimum
height=1cm,text centered, draw=black, fill=blue!10!white, dashed]
\tikzstyle{arrow} = [thick,->,>=stealth]
\usetikzlibrary{calc, quotes}

\tikzstyle{fblock} = [rectangle, draw, fill=gray!20, 
text width=8em, text centered, rounded corners, minimum height=4em, minimum width = 8em]
\tikzstyle{line} = [draw, -latex']

\usepackage{algorithmicx}
\usepackage{algorithm}
\usepackage[noend]{algpseudocode}

\usepackage{fontawesome}

\usepackage{soul}
\usepackage{comment}
\usepackage[]{url}
\usepackage[font=small,hypcap=false]{caption}
\usepackage[normalem]{ulem}
\useunder{\uline}{\ul}{}
\usepackage{makecell}

\usepackage[colorlinks=true, linkcolor=blue, citecolor=black, urlcolor=cyan, filecolor=black, pagebackref=false, hypertexnames=false]{hyperref}
\usepackage{cleveref}
\crefname{problem}{Problem}{Problems}
\crefname{example}{Example}{Examples}
\crefname{section}{Sec.}{Secs.}
\Crefname{section}{Section}{Sections}
\Crefname{table}{Table}{Tables}
\crefname{table}{Table}{Tabs.}
\crefname{figure}{Fig.}{Figs.}
\crefname{algorithm}{Algorithm}{Algorithms}
\crefname{remark}{Remark}{Remarks}
\crefname{theorem}{Theorem}{Theorems}
\crefname{proposition}{Proposition}{Propositions}
\crefname{lemma}{Lemma}{Lemmas}
\crefname{corollary}{Corollary}{Corollaries}
\crefname{assumption}{Assumption}{Assumptions}
\crefname{definition}{Definition}{Definitions}
\crefname{appendix}{Appendix}{Appendices}
\Crefname{appendix}{Appendix}{Appendices}

\newcommand{\AlgName}{{\textsc{SERF}}\xspace}

\newcommand{\Real}{\mathbb{R}}

\DeclareMathOperator*{\argmin}{arg\,min}

\DeclareMathOperator{\SOd}{SO}

\DeclareMathOperator{\SE}{SE}

\newcommand{\bmat}{\begin{bmatrix}}
\newcommand{\emat}{\end{bmatrix}}

\newcommand{\etal}{\emph{et~al.}\xspace}

\newcommand{\eg}{\emph{e.g.}\xspace}

\providecommand{\method}[1]{{\small \textsf{#1}}\xspace}

\newcommand{\myParagraph}[1]{{\bf #1.}\xspace}

\providecommand{\optional}[1]{{}}

\providecommand{\techreport}[1]{{}}  

\newcolumntype{Y}{>{\centering\arraybackslash}X}

\newcommand{\calD}{\mathcal{D}}

\newcommand{\calF}{\mathcal{F}}

\newcommand{\calI}{\mathcal{I}}

\newcommand{\calL}{\mathcal{L}}

\newcommand{\calN}{\mathcal{N}}

\newcommand{\calP}{\mathcal{P}}

\newcommand{\calT}{\mathcal{T}}

\newcommand{\calX}{\mathcal{X}}
\newcommand{\calY}{\mathcal{Y}}

\title{SERF: Spatiotemporal Environment and Robot Feature Map for Long-Horizon Mobile Manipulation}

\author{%
{
Sunghwan Kim$^{1*}$ \quad
Byeonghyun Pak$^{2*}$ \quad
Kehan Long$^{1}$ \quad
Yulun Tian$^{3}$ \quad
Nikolay Atanasov$^{1}$}\\[1.2ex]
{\normalfont\small $^{1}$UC San Diego \quad
$^{2}$Agency for Defense Development \quad
$^{3}$University of Michigan}\\
}

\hypersetup{
  pdftitle={SERF: Spatiotemporal Environment and Robot Feature Map for Long-Horizon Mobile Manipulation},
  pdfauthor={Sunghwan Kim, Byeonghyun Pak, Kehan Long, Yulun Tian, Nikolay Atanasov}
}

\begin{document}
\maketitle
\begingroup
\renewcommand{\thefootnote}{\fnsymbol{footnote}}
\footnotetext[1]{\,These authors contributed equally. Correspondence to: \texttt{suk063@ucsd.edu}, \texttt{bhpak@umd.edu}}
\endgroup
\vspace{-0.3in} 

\begin{figure}[H]
    \centering
    \includegraphics[width=\textwidth]{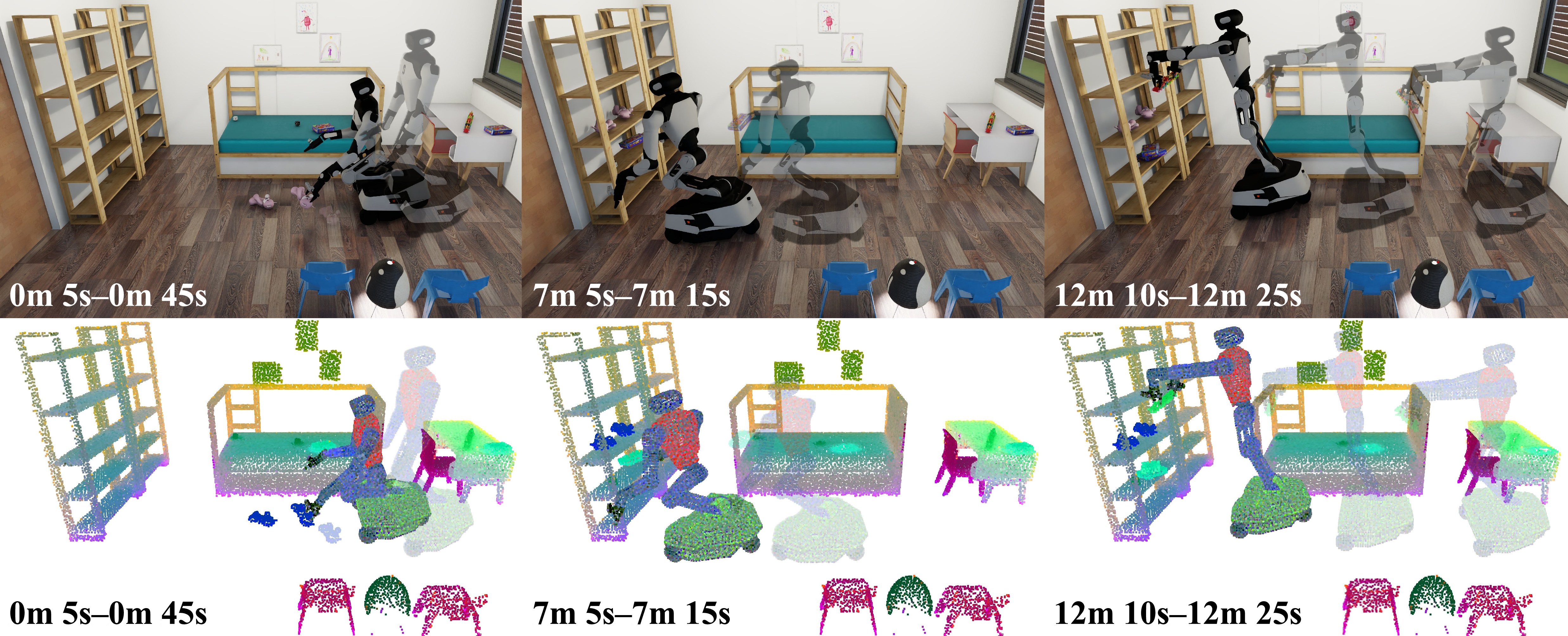}
    \caption{\textbf{Top:} A mobile manipulator performs a long-horizon task consisting of multiple subgoals.
    \textbf{Bottom:} A spatiotemporal feature map represents the evolving environment and the robot body in a shared latent space, visualized via PCA. The map is updated online from egocentric observations and proprioceptive state. Video results are available at the project website: \href{https://existentialrobotics.org/serf/}{\texttt{https://existentialrobotics.org/serf/}}.}
    \label{fig:teaser}
\end{figure}

\begin{abstract}
Long-horizon robot mobile manipulation requires continual reasoning about localization, environment changes, and task progress, all of which are challenging to infer from image observations alone.
In this paper, we show that conditioning a mobile manipulation policy on a spatiotemporal feature map improves reasoning over long horizons. 
The map represents the environment and the articulated robot body as neural points in a shared latent space and is updated online from egocentric observations and proprioceptive state.
We update the environment neural points using object-level rigid tracking and the robot neural points using forward kinematics.
We use our spatiotemporal environment and robot feature (\AlgName) map as a state input to a vision-language-action (VLA) model by extracting map tokens from multiple reference frames and spatial scales, providing the policy with both local and global context.
We demonstrate \AlgName on BEHAVIOR-1K, a benchmark for long-horizon mobile manipulation in household environments.
Experiments show that the \AlgName VLA policy outperforms image-only baselines, reaches subgoals faster by following more direct trajectories, improves robustness to scene-configuration shifts, and recovers from object-drop failures.
\end{abstract}
\section{Introduction}
\label{sec:intro}
Recent advances in robot learning have enabled impressive manipulation capabilities in short-horizon tabletop settings.
Extending these successes to long-horizon mobile manipulation in large environments remains an open challenge.
Consider a mobile manipulator tasked with tidying up a child's room (\cref{fig:teaser}).
It must collect scattered toys across the room, carry them around furniture, and place them on the shelves of a bookcase.
Such tasks require the robot to continually answer three coupled questions: \emph{Where am I?} \emph{What has changed around me?} \emph{How far along am I in my task?}
Answering these questions jointly requires \emph{coherent spatiotemporal reasoning}: maintaining a unified understanding of the robot's motion and evolving environment over long horizons.


Vision-language-action (VLA) models have shown promising results in robot manipulation by mapping visual observations and language instructions directly to robot actions \cite{zitkovich2023rt2, black2024vision, black2025pi05, kim24openvla, team2025gemini, nvidia2025gr00t, intelligence2025pi06}.
However, most VLA policies encode spatial and temporal context without an explicit mechanism for maintaining information beyond the current observation.
This limitation becomes critical in mobile manipulation, where action decisions depend on long-term context that evolves with the robot's interactions.

Recent work explores persistent memory formulations to extend the information available to embodied agents beyond their current observations.
One line of work builds modular mobile manipulation systems that integrate perception, navigation, planning, and manipulation \cite{liu2024ok, chen2026owmm}, in which persistent scene representations such as dynamic open-vocabulary 3D scene graphs \cite{yan2025dynamic}, online spatio-semantic object memories \cite{liu2024dynamem}, task-relevant scene-graph abstractions \cite{mohammadi2025more}, and predictive world models \cite{bar2024navigationworldmodels} provide structured spatial grounding for long-horizon reasoning.
These approaches provide explicit, interpretable scene memory, but they often abstract the scene into objects, symbols, or high-level planning states.
Such abstractions support semantic reasoning, but may discard dense geometry and robot--environment interaction cues needed for continuous visuomotor control.

A second line of work embeds memory directly into policy architectures in the form of episodic memories, history tokens, retrieval buffers, or language-indexed long-term memory \cite{sridhar2025memer, torne2026mem, xin2025echovla}. These approaches extend temporal context, but often leave the spatial grounding implicit: the policy must infer which past observations are relevant to the current robot state and subgoal.
Some approaches explicitly consider spatial grounding by conditioning policies on 3D feature maps distilled from vision foundation models (VFMs), including semantic 3D reconstruction maps \cite{steiner2025mindmap}, latent 3D feature maps \cite{kim2025seeing}, and geometry-grounded world models \cite{zhen20243dvla}.
These maps provide spatial memory of object and goal locations.
However, existing feature-map approaches typically represent only the static portion of the environment without support for updates as the robot interacts with the environment.
These methods also omit the robot body from the memory, requiring robot--environment relationships to be inferred indirectly from separate visual or proprioceptive inputs.
These limitations motivate the development of a spatiotemporal memory that evolves with robot interaction, explicitly represents robot--environment spatial relationships, and provides structured context for neural policies.


In this paper, we introduce a spatiotemporal (4D) feature mapping formulation that captures both the environment and the robot in a shared latent feature representation, which evolves over time as the robot interacts with the environment (see \cref{fig:teaser}). 
This shared robot--environment feature map provides persistent, spatially grounded memory for policy learning, supporting two forms of reasoning.
(i)~\emph{Allocentric reasoning}: the map provides persistent spatial memory for situating the robot in the global scene and tracking object and goal locations over time.
(ii)~\emph{Egocentric reasoning}: embedding the robot body in the map exposes robot--environment relationships, such as distance, orientation, and reachability, which support spatially grounded decision making.

We build the spatiotemporal feature map using \emph{neural points}~\cite{aliev2020neural, pan2024pin}, which are 3D points with learnable latent features trained to reconstruct dense VFM embeddings (\eg, DINOv3~\cite{simeoni2025dinov3}).
To track changes, we construct 3D keypoint correspondences between consecutive observations, estimate an object-level $\SE(3)$ transform, and update the corresponding points.
We extend the neural points to the articulated robot body by sampling surface points from a robot kinematic model (\eg, a URDF) and positioning them via forward kinematics at each step.
We train the environment and robot neural points with a shared decoder, encouraging both sets to lie in a common latent space.
Notably, we construct and maintain the map using egocentric observations and proprioceptive state (\cref{sec:4d_latent_feature_mapping}). We refer to this representation as the \underline{S}patiotemporal \underline{E}nvironment and \underline{R}obot \underline{F}eature (\AlgName) map.

To demonstrate the utility of \AlgName for decision making, we develop a map-conditioned VLA policy that uses the \AlgName map as a structured state input. We tokenize the map across multiple reference frames and spatial scales, including end-effector and robot-base frames, to provide both local and global context to the VLA (\cref{sec:manipulation}). We evaluate \AlgName on BEHAVIOR-1K~\cite{li2022behavior}, a benchmark of long-horizon household bimanual mobile manipulation tasks involving navigation and multi-object rearrangement. Experiments show that the \AlgName VLA policy (i) outperforms image-only VLA baselines, improving average task progress across three BEHAVIOR-1K tasks from $44.0\%$ to $58.7\%$, (ii) reaches subgoals faster by following more direct trajectories, (iii) improves robustness to scene-configuration shifts, \eg, increasing task progress in unvisited regions from $28.0\%$ to $51.0\%$, and (iv) recovers from execution failures by retrieving dropped objects (\cref{sec:evaluation}).



In summary, we make the following contributions.
\begin{itemize}[nosep,leftmargin=*]
    \item We introduce a spatiotemporal environment and robot feature (\AlgName) map that represents the robot body and the evolving environment in a shared latent feature space.
    

    \item We design a map-conditioned VLA policy that uses \AlgName tokens across multiple reference frames and spatial scales, providing both local and global spatial context for action prediction.
    
    \item We show that the \AlgName VLA policy outperforms image-only VLA baselines on long-horizon mobile manipulation, improving task progress, trajectory efficiency, scene-configuration generalization, and failure recovery.
\end{itemize}

\section{Problem Formulation}
\label{sec:problem}

\begin{figure}[t]
  \centering
  \begin{minipage}[t]{0.485\linewidth}
    \centering
    \includegraphics[height=4.3cm, keepaspectratio]{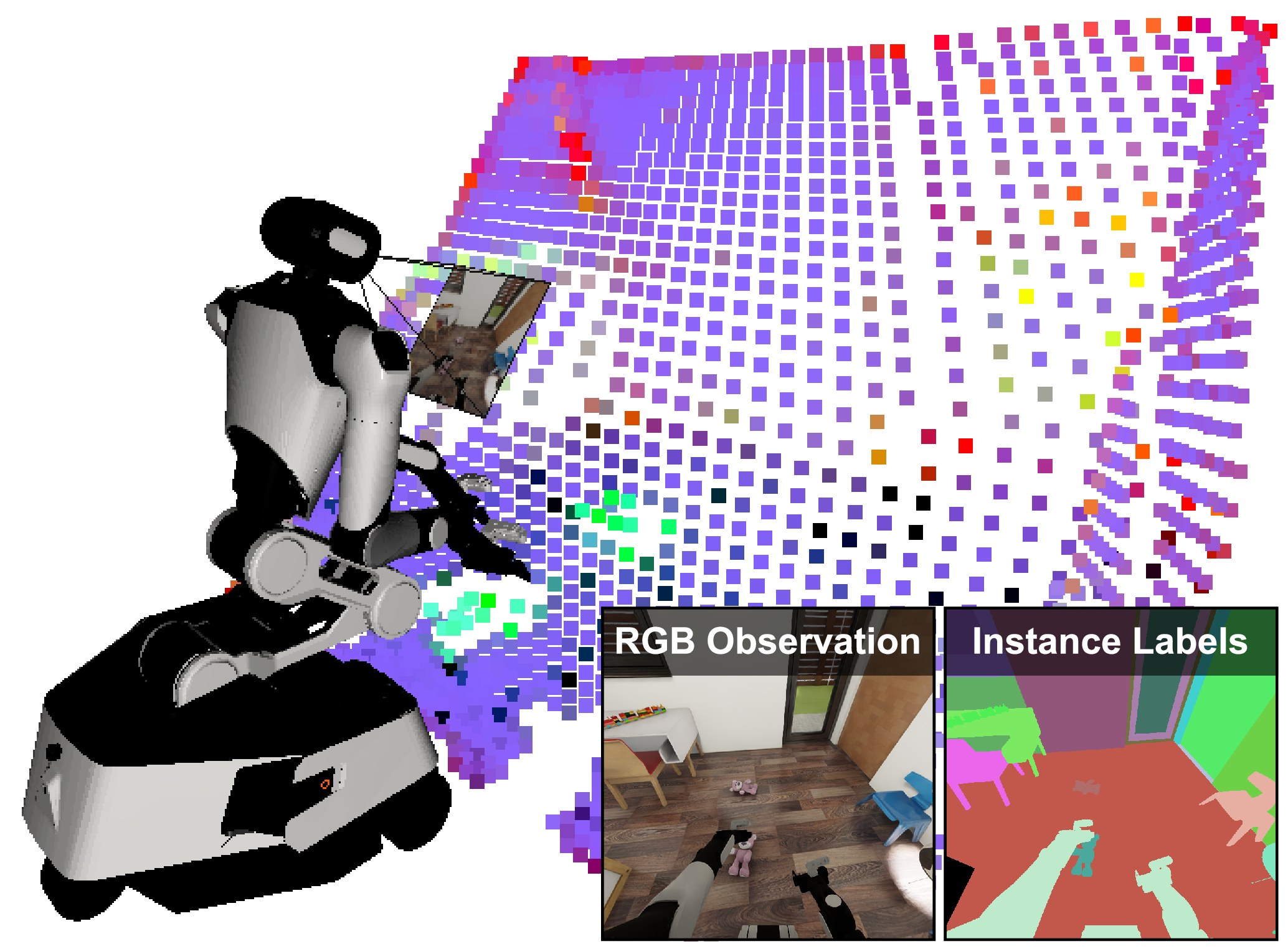}
    \caption{
      Per-patch VFM embeddings from robot observations are back-projected into 3D.
      }
    \label{fig:featurized_point_cloud}
  \end{minipage}%
  \hfill%
  \begin{minipage}[t]{0.485\linewidth}
    \centering
    \includegraphics[height=4.3cm, keepaspectratio]{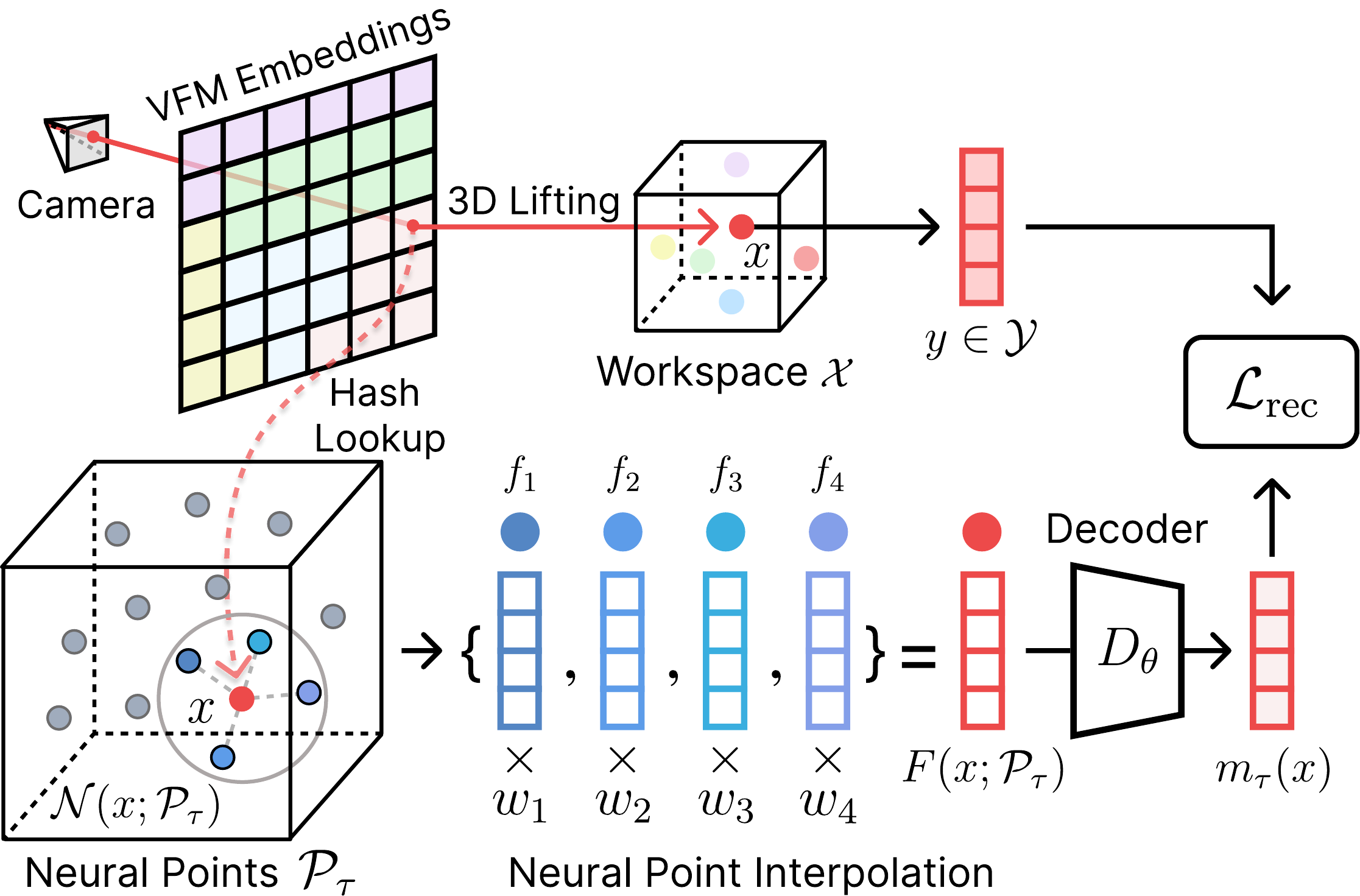}
    \caption{
      Neural point features are interpolated and decoded to reconstruct per-patch VFM embeddings.
      }
    \label{fig:mapping_framework}
  \end{minipage}
  \vspace{-10pt}
\end{figure}

We consider a mobile manipulator operating in a workspace $\calX \subseteq \Real^3$. At each time step $\tau$, the robot has access to its proprioceptive state $s_\tau$, including the robot base pose and joint angles, as well as RGB-D observations $o_\tau=\{(I_\tau,Z_\tau)\}$ from its head and wrist cameras, where $I_\tau$ is an RGB image and $Z_\tau$ is a depth image. 
Each RGB image $I_\tau$ is associated with instance segmentation labels $C_\tau$, which assign each pixel an integer instance ID.
The labels can be obtained either from a segmentation model (\eg, SAM 2~\cite{ravi2024sam}) or directly from a simulation environment.
We use a pretrained VFM (\eg, DINOv3~\cite{simeoni2025dinov3}) to extract per-patch embeddings $Y_\tau \in \calY^{H \times W}$, where $\calY \subseteq \Real^k$ is the VFM embedding space.
Using depth information and the camera pose, the image patch centroids are back-projected into 3D coordinates and paired with the VFM embeddings, yielding coordinate--embedding pairs $(x,y)\in\calX\times\calY$; see \cref{fig:featurized_point_cloud}.
The objective of the robot is to execute a task specified by a natural language command $\ell$. Let $e_\ell$ denote a task embedding of the language command. We assume access to an expert demonstration dataset $\calT$ of observation, state, action, language tuples $(o_\tau,s_\tau,a_\tau,\ell)$ collected from multiple expert trajectories.


We aim to learn a policy $\pi_\phi(a_\tau \mid o_{1:\tau}, s_{1:\tau}, e_\ell)$ for the mobile manipulator from the expert demonstrations $\calT$. Instead of conditioning the policy $\pi_\phi$ on the full observation--state history, we summarize this history in a spatiotemporal (4D) feature map $m_\tau$ of the evolving workspace and condition the policy on this map.
This formulation leads to two learning problems.
First, we learn a feature map $m_\tau\colon \calX \to \calY$ that summarizes $(o_{1:\tau},s_{1:\tau})$ by reconstructing VFM embeddings at workspace coordinates (\cref{prob:map}).
This map is incrementally updated from $m_{\tau-1}$ using the current observation and state $(o_\tau,s_\tau)$.
Second, we learn a policy $\pi_\phi(a_\tau \mid m_\tau, o_\tau, s_\tau, e_\ell)$ that uses this persistent map, together with the current observation, state, and task embedding, to imitate expert actions (\cref{prob:policy}).
For the mapping problem, let $\calD_\tau \subseteq \calX \times \calY$ denote the coordinate--embedding pairs available by time $\tau$, derived from onboard observations.
We write $\calD_0$ for the prior reconstruction dataset constructed from pre-execution observations using the same coordinate--embedding procedure.



\begin{problem}[Spatiotemporal Feature Mapping]
    \label{prob:map}
    Learn map parameters $\Theta$ by minimizing
    $\calL_{map}(\Theta)=
    \mathbb{E}_{\tau,\,(x,y)\sim\calD_\tau}
    \!\left[\calL(m_\tau(x;\Theta),y)\right]$,
    where $\tau\in\{0,\ldots,T\}$ and $\calL\colon \calY \times \calY \to \Real_{\geq 0}$ is a reconstruction loss.
\end{problem}

\begin{problem}[Map-Conditioned Behavior Cloning]
    \label{prob:policy}
    Given spatiotemporal feature maps $m_\tau$ for demonstrations in $\calT$, learn $\pi_\phi$ by minimizing
    $\calL_{bc}(\phi)=
    -\mathbb{E}_{(o_\tau,s_\tau,a_\tau,\ell)\sim\calT}
    \!\left[\log \pi_\phi(a_\tau \mid m_\tau,o_\tau,s_\tau,e_\ell)\right]$.
\end{problem}

The key difference from standard behavior cloning is that our policy uses the map $m_\tau$ as an explicit state input.
In the rest of the paper, we formulate a representation and learning approach for spatiotemporal feature mapping (\cref{sec:4d_latent_feature_mapping}), design a policy that conditions action predictions on the map (\cref{sec:manipulation}), and evaluate the approach on long-horizon mobile manipulation tasks (\cref{sec:evaluation}).







\section{Spatiotemporal Feature Mapping}
\label{sec:4d_latent_feature_mapping}

We parameterize the feature map $m_\tau$ using neural points in 3D space, where each point carries a learnable latent feature.
The map contains separate neural point sets for the environment and the robot body.
The latent features and a shared decoder are optimized offline to reconstruct VFM embeddings with a cosine-similarity objective.
Additional contrastive losses align features within the same category or part and separate features across categories or parts, encouraging semantic structure in the latent space.
During task execution, the learned features remain fixed, and only the point coordinates are updated.
We move the environment points by estimating object-level $\SE(3)$ transforms from 3D keypoint correspondences, and move the robot points via forward kinematics from the current proprioceptive state.

\myParagraph{Neural Point Representation}
We define the map $m_\tau$ at time $\tau$ as a function $\calX \to \calY$ mapping query points $x$ in the workspace to the VFM embedding space $\calY$. We represent the map using a set of neural points $\calP_\tau$ in the workspace.
To query the map at location $x$, we interpolate latent features from nearby points in $\calP_\tau$ and use a neural network decoder $D_\theta$ to map the interpolated feature to a VFM embedding (see \cref{fig:mapping_framework}).
We denote the neural point set as $\calP_\tau = \{(p_{i,\tau}, f_i, c_i)\}_{i=1}^{N}$, where $p_{i,\tau}\in\calX$ is the world-frame position of point $i$ at time $\tau$, $f_i \in \calF$ is a latent feature vector, $\calF \subseteq \Real^c$ denotes the latent feature space, and $c_i$ is the object instance label of point $i$.
Because the neural points have explicit coordinates, they can be repositioned under rigid-body motion, making them a good primitive for dynamic scenes.

The decoder $D_\theta\colon \calF \to \calY$ is an MLP that maps features to the VFM embedding space.
Given $\calP_\tau$, we query the latent feature at a location $x \in \calX$ by gathering candidate points within a ball query, selecting the $K$ nearest neighbors $\calN(x;\calP_\tau)$, and interpolating their latent features:
\begin{equation}\label{eq:latent_feature_query}
F(x;\calP_\tau) = \sum_{i\in\calN(x;\calP_\tau)} w_i(x;\calP_\tau)\, f_i,\quad w_i(x;\calP_\tau)=\mathrm{softmax}\bigl(-\|x-p_{i,\tau}\|/\sigma\bigr),\quad \sigma>0.
\end{equation}
Together, $\calP_\tau$ and $D_\theta$ define the map function as $m_\tau(x;\Theta) = D_\theta(F(x;\calP_\tau))$.
The latent features $\{f_i\}$ and decoder parameters $\theta$ form the learnable map parameters $\Theta$ in \cref{prob:map}. We maintain neural points for both the environment and the robot body, $\calP_\tau = \calP^e_\tau \cup \calP^r_\tau$, described next.



\myParagraph{Environment Points}
We lift VFM patch locations from robot observations into 3D, voxelize the resulting points, and register one environment neural point for each newly occupied voxel in a spatial hash table~\cite{pan2024pin}.
We assign each point an instance label from segmentation and randomly initialize its latent feature. The environment points $\calP^e_\tau = \{(p^e_{i,\tau}, f^e_i, c^e_i)\}_{i=1}^{N_e}$ are updated by having their positions $p^e_{i,\tau}$ track moving objects (discussed below), while the features $f^e_i$ and labels $c^e_i$ remain fixed. Environment queries use the interpolation rule in \eqref{eq:latent_feature_query}. 
We learn the latent features of the initial points $\calP^e_0$ before execution.
For each environment reconstruction sample $(x,y)$, we optimize these features using the cosine-similarity reconstruction loss $\calL_{\text{rec}} = 1 - \mathrm{sim}(D_\theta(F(x;\calP^e_0)), y)$.


\myParagraph{Robot Points}
We represent the robot body with a separate set of neural points.
Given a robot kinematic model (\eg, a URDF), we sample surface points from the robot link meshes and store each point in its corresponding local link frame, yielding $\calP^r = \{(u_j, f^r_j, c^r, l_j)\}_{j=1}^{N_r}$.
Here, $u_j \in \Real^3$ is a surface point in the local frame of its associated link $l_j$, $f^r_j$ is its latent feature, and $c^r$ is a shared robot-body label that distinguishes robot points from environment points.
We initialize the robot latent features randomly.
Given a robot state $s$, we use forward kinematics to obtain the world-frame robot point set $\calP^r(s) = \{(p^r_j(s), f^r_j, c^r)\}_{j=1}^{N_r}$.
Here, $p^r_j(s) = R_{l_j}(s)u_j + t_{l_j}(s)$, where $(R_{l_j}(s), t_{l_j}(s)) \in \SE(3)$ is the world-frame pose of link $l_j$.
Thus, robot point positions are determined by fixed link-frame coordinates and the robot state.
For robot queries at state $s$, we apply the same interpolation rule to $\calP^r(s)$: $F^r(x;s) \coloneqq F(x;\calP^r(s))$.
The corresponding robot map prediction is $D_\theta(F^r(x;s))$.
For each robot reconstruction sample $(x,y)$ generated at robot state $s$ (\cref{sec:coordinate_feature_pairs}), we apply the same cosine-similarity reconstruction objective to the state-conditioned prediction: $\calL_{\text{rec}} = 1 - \mathrm{sim}(D_\theta(F^r(x;s)), y)$.
This state-conditioned supervision encourages each robot point to maintain a consistent semantic identity across robot configurations.

\myParagraph{Latent Feature Learning}
During offline map learning, we query the environment and robot neural point sets separately, while jointly optimizing their latent features and the shared decoder $D_\theta$ with reconstruction supervision from both environment and robot samples.
The decoder encourages environment and robot features to lie in a shared latent feature space.
We provide additional map representation and training details in \cref{sec:map_representation_details}.

In addition to the reconstruction objective, we optimize the latent features with contrastive losses:
$\calL_{\text{map}} = \calL_{\text{rec}} + \lambda_{\text{inter}}\calL_{\text{inter}} + \lambda_{\text{intra}}\calL_{\text{intra}}$.
The inter-category objective $\calL_{\text{inter}}$ pulls same-category features together and pushes different-category features apart across training scenes. 
The intra-instance objective $\calL_{\text{intra}}$ pulls features from the same object part together and separates features from different parts of the same instance, inducing part-level structure~\cite{kim2024garfield}.
Additional details and visualizations of the learned feature structure are provided in \cref{sec:contrastive_losses}.

\myParagraph{Map Updates}
At the start of execution, the initial map $m_0$ is defined by the environment points $\calP^e_0$, robot points $\calP^r(s_0)$, and the decoder $D_\theta$. For $\tau \geq 1$, we update only the point positions, while keeping the latent features, instance labels, and decoder fixed. 
For environment updates, we model each object instance as a rigid body and group environment neural points by their instance labels.
Let $\calI \subset \{1,\ldots,N_e\}$ be the index set of neural points belonging to a given instance.
For each instance, we track 2D keypoints~\cite{shi1994good,karaev2024cotracker} between consecutive observations $(\tau-1,\tau)$ and lift the tracks to 3D.
We estimate the object motion 
by solving $(\hat{R},\hat{t}) = \argmin_{R\in\SOd(3),\,t\in\Real^3} \sum_k \|q^\tau_k - (R q^{\tau-1}_k + t)\|^2$, where the sum is over lifted 3D keypoint correspondences $(q^{\tau-1}_k, q^\tau_k)$.
In practice, we initialize this estimate with Fast Global Registration (FGR)~\cite{zhou2016fast} and refine it with Iterative Closest Point (ICP)~\cite{besl1992method} against the observed point cloud at time $\tau$.
We update all environment neural points of the instance as $p^e_{i,\tau} = \hat{R}p^e_{i,\tau-1} + \hat{t}$ for all $i\in\calI$.

For robot updates, after observing the current proprioceptive state $s_\tau$, we apply forward kinematics to compute each link transform.
For a robot point $j$ stored at link-frame coordinate $u_j$ on link $l_j$, its world-frame position at time $\tau$ is $p^r_{j,\tau} = R_{l_j}(s_\tau)u_j + t_{l_j}(s_\tau)$, where $(R_{l_j}(s_\tau), t_{l_j}(s_\tau)) \in \SE(3)$ is the world-frame transform of link $l_j$.
This yields the world-frame robot point set $\calP^r_\tau \coloneqq \calP^r(s_\tau) = \{(p^r_{j,\tau}, f^r_j, c^r)\}_{j=1}^{N_r}$.
Together, $\calP^e_\tau$ and $\calP^r_\tau$ form the full scene point set $\calP_\tau = \calP^e_\tau \cup \calP^r_\tau$, which determines the map $m_\tau$.
Additional details on the map updates are provided in \cref{sec:map_update_details}.

\section{Map-Conditioned VLA Policy}
\label{sec:manipulation}

\begin{figure*}[t]
  \centering
  \includegraphics[width=\linewidth]{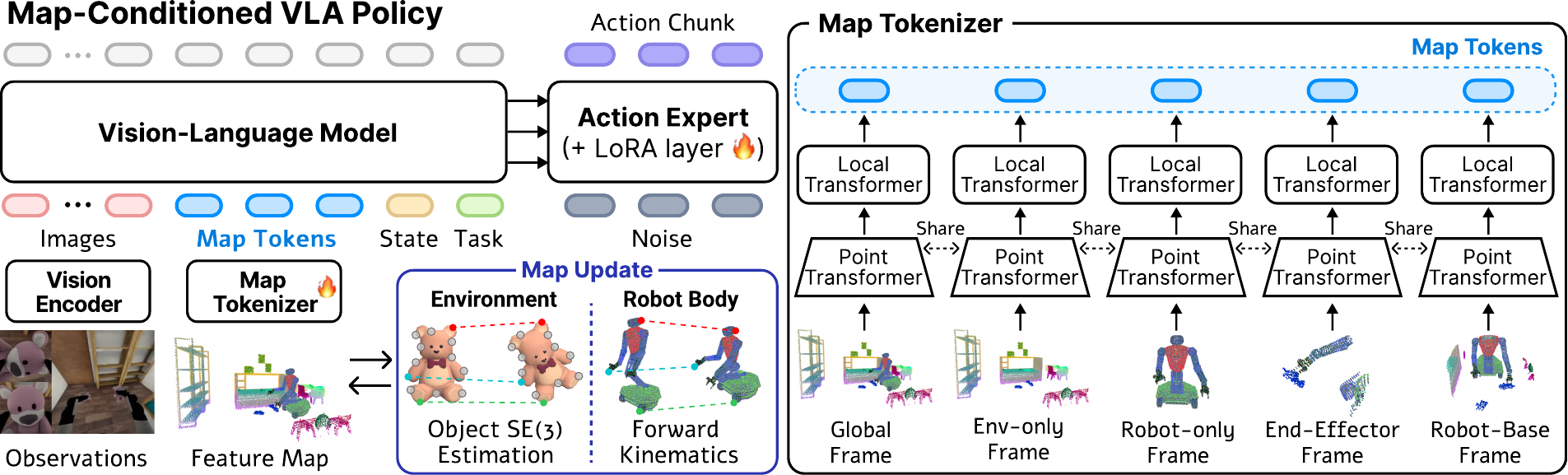}
  \caption{
    \textbf{Overview of map-conditioned VLA policy.}
    A map tokenizer produces map tokens across multiple reference frames and spatial scales.
    The VLA model is conditioned on these map tokens, along with image observations, the task embedding, and proprioceptive state, to predict actions.
 }
  \label{fig:vla}
  \vspace{-10pt}
\end{figure*}

To obtain a mobile manipulation policy $\pi_\phi$, we condition a VLA model on the spatiotemporal feature map as an explicit state input.
We design a map tokenizer that extracts tokens from the map across multiple reference frames and spatial scales, providing the policy with both local and global context.  
\Cref{fig:vla} illustrates the map tokenizer and the map-conditioned VLA policy architecture.

\myParagraph{Map Tokenizer}
 We tokenize a filtered subset of the neural point set $\calP_\tau$ to produce map tokens.
Before tokenization, we use stored instance labels to retain points belonging to the robot body and objects referenced by the task specification (\eg, BDDL~\cite{srivastava2021behavior}), while discarding background and structural points. 
A Point Transformer~\cite{zhao2021point} backbone encodes the filtered point set into point features.
Eight parallel heads then select spatial subsets of these features via ball queries or mask-based selection, each producing one map token.
These heads are organized into five groups:
(i)~three \emph{robot-base} tokens at increasing radii ($1$\,m, $2$\,m, $4$\,m), capturing local context at multiple scales;
(ii)~two \emph{end-effector} tokens extracted by $0.5$\,m radius ball queries around the left and right grippers to support grasp reasoning;
(iii)~a \emph{robot-only} token summarizing the robot body configuration;
(iv)~an \emph{environment-only} token summarizing the current environment state; and
(v)~a \emph{global} token aggregating all points for scene-level reasoning.
Each branch applies Point Transformer layers followed by attention pooling to produce a single map token. To reduce overfitting, we omit absolute coordinates from the token features and use only Point Transformer relative positional encodings. We provide an ablation of the five map-token groups in \cref{sec:map_token_ablation}.


\myParagraph{Map-Conditioned Policy Learning}
We construct the map-conditioned policy by providing map tokens as additional input tokens to a VLA model.
We build on $\pi_{0.5}$~\cite{black2025pi05}, a VLA whose VLM backbone produces prefix tokens from observations, proprioception, and task information, while its action expert predicts robot actions from these tokens. 
Following~\cite{larchenko2025task}, we use a learnable task embedding $e_\ell$ to represent the task specification. 
RGB observations are encoded by the VLA vision encoder~\cite{zhai2023sigmoid}, while the proprioceptive state $s_\tau$ is discretized and embedded as state tokens $e_s$.
Map tokens are projected into the VLA token space to form
$e_m = [\tilde{z}_1, \ldots, \tilde{z}_8]$.
We then concatenate the image features, map tokens, state tokens, and task embedding:
$h_\tau = \operatorname{Concat}(E(o_\tau), e_m, e_s, e_\ell)$,
where $E$ denotes the vision encoder.
We optimize the policy $\pi_\phi$ in \cref{prob:policy} with a conditional flow-matching objective. 
Given an expert action chunk $A_\tau=a_{\tau:\tau+H}$ and standard Gaussian noise $\epsilon$, we define $A_\tau^\alpha=\alpha A_\tau+(1-\alpha)\epsilon$ with $\alpha\in[0,1]$.
We train the action expert $v_\psi$ to predict the target velocity $u_\tau=A_\tau-\epsilon$ by minimizing $\mathcal{L}_{\mathrm{action}}=\mathbb{E}_{(h_\tau,A_\tau)\sim\mathcal{T},\,\alpha,\epsilon}\left[\|v_\psi(A_\tau^\alpha,h_\tau,\alpha)-u_\tau\|_2^2\right]$.
At inference time, we initialize $\hat{A}_\tau^0=\epsilon$ and integrate the velocity field $v_\psi(\hat{A}_\tau^\alpha,h_\tau,\alpha)$ from $\alpha=0$ to $\alpha=1$ to obtain $\hat{a}_{\tau:\tau+H}=\hat{A}_\tau^1$.
To preserve pretrained knowledge, we freeze the VLM backbone and vision encoder, and insert LoRA layers~\cite{hu2022lora} into the action expert.
The map tokenizer, projection layer, and LoRA parameters constitute the policy parameters $\phi$ in \cref{prob:policy}.
We provide additional implementation details for the map tokenizer and VLA policy in \cref{sec:vla_policy_details}.

\section{Long-Horizon Mobile Manipulation}
\label{sec:evaluation}

This section evaluates whether the \AlgName VLA policy improves long-horizon mobile manipulation performance over image-only VLA baselines by providing persistent spatiotemporal memory.
We evaluate on three BEHAVIOR-1K household tasks~\cite{li2022behavior}, where the robot must navigate large workspaces and complete multi-object rearrangement.
The full \AlgName policy achieves the highest task progress across all three evaluated tasks, outperforming image-only policies.
We also evaluate whether \AlgName supports scene-configuration generalization and failure recovery.


\myParagraph{Implementation Details}
To simplify data association, we use privileged instance labels from the simulator rather than deriving them from RGB images.
\Cref{sec:map_representation_details,sec:contrastive_losses,sec:map_update_details,sec:vla_policy_details} provide implementation details about the map representation, contrastive objectives, map updates, and the VLA policy.

\myParagraph{Benchmark}
BEHAVIOR-1K~\cite{li2022behavior} is a benchmark of household bimanual mobile manipulation tasks in OmniGibson.
We evaluate three tasks: Task 21 (\emph{Collecting Children's Toys}), Task 22 (\emph{Putting Shoes On Rack}), and Task 26 (\emph{Assembling Gift Baskets}).
For each task, we report task progress (\%) across 20 evaluation configurations, measured as the fraction of completed subgoals in the BDDL task specification.
Additional benchmark details are provided in \cref{sec:behavior_1k_details}.

\begin{figure*}[t]
  \centering
  \includegraphics[width=\textwidth]{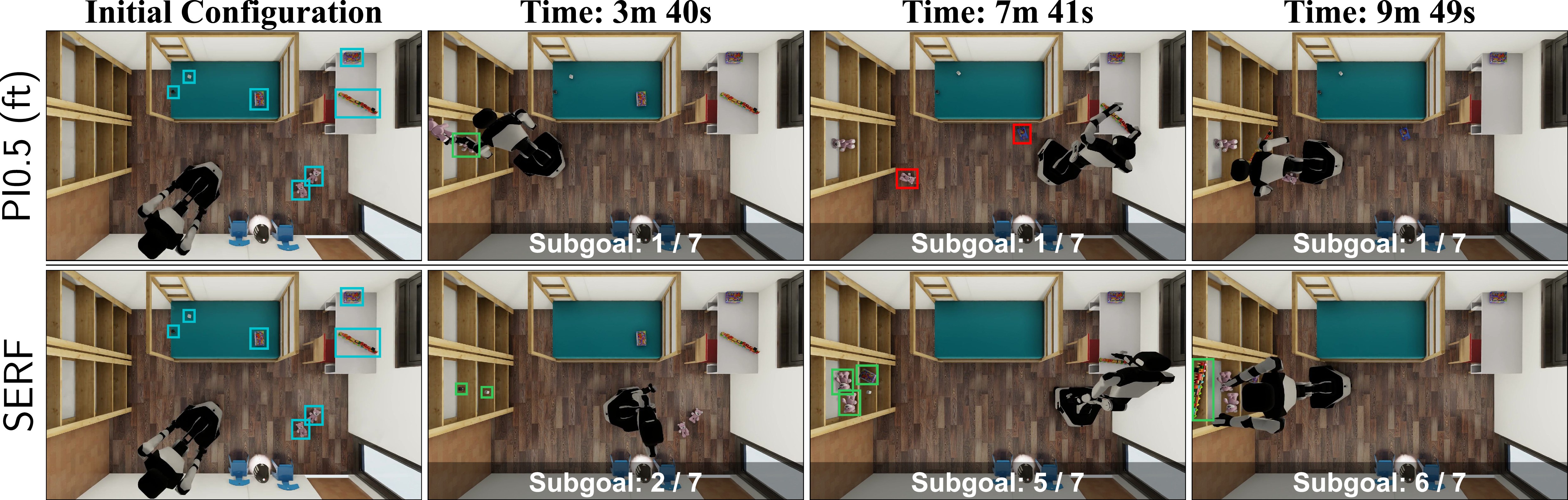}
  \caption{
  \textbf{Qualitative comparison on long-horizon mobile manipulation.}
  Map-conditioned \method{SERF} takes more direct trajectories than image-only \method{PI0.5~(ft)}, reaches subgoals faster, and achieves higher task progress.
  }
  \label{fig:qualitative_comparison}
  \vspace{-15pt}
\end{figure*}


\myParagraph{Baselines}
We compare five VLA policy variants that differ in whether and how they use map information.
(i)~\method{PI0.5 (pre)} uses the $\pi_{0.5}$ checkpoint from \cite{larchenko2025task}, obtained by fine-tuning the original $\pi_{0.5}$ model on all 50 BEHAVIOR-1K tasks. We initialize all other variants from this checkpoint.
(ii)~\method{PI0.5~(ft)} is an image-only VLA policy fine-tuned separately on each target task.
(iii)~\method{SBP} augments the VLA policy with a static 3D feature map~\cite{kim2025seeing}.
(iv)~\method{SERF~(env)} is a VLA policy variant that uses the \AlgName map but excludes robot neural points.
(v)~\method{SERF} denotes the full \AlgName policy, which uses both environment and robot neural points.
For a fair comparison, all variants share the same $\pi_{0.5}$ backbone and training setup; they vary only in their map-token inputs.
Because \method{SBP} and \method{SERF~(env)} contain only environment points, they use six rather than eight map tokens: the robot-only token is undefined, and the global token would duplicate the environment-only token.

\myParagraph{Results}
\Cref{tab:quantitative_result} summarizes the quantitative evaluation results.
The full \AlgName policy achieves the highest task progress on all three evaluated tasks.
Map-conditioned policies achieve higher mean task progress than image-only baselines, supporting the benefit of explicit spatial memory.
Among map variants, \method{SERF~(env)} improves over the static \method{SBP} baseline on average, indicating the benefit of temporal map updates.
Adding robot neural points yields further average gains, suggesting that explicit robot–environment spatial relationships can support egocentric spatial reasoning.
Appendix~\ref{sec:image_history} presents an additional comparison with an image-history baseline~\cite{koo2026hamlet}, showing that \method{SERF} consistently outperforms this baseline.
We also provide per-step task progress curves in \cref{sec:time_to_subgoal}.
\Cref{fig:qualitative_comparison} compares representative rollouts from \method{PI0.5~(ft)} and \method{SERF} on Task~21.
Additional qualitative comparisons are provided in \cref{sec:more_qualitative_results}.
\method{SERF} follows more direct trajectories and reaches subgoals faster, whereas \method{PI0.5~(ft)} lacks explicit spatial memory and often stalls when task-relevant objects leave the current field of view.



\myParagraph{Scene-Configuration Generalization}
We find that the \AlgName policy can generalize to new scene configurations that differ from the demonstration layouts.
We evaluate this capability by comparing \method{SERF} against the image-only \method{PI0.5~(ft)} and the map-conditioned \method{SBP} and \method{SERF~(env)} under shifted object and goal configurations at test time.
We consider three test-time out-of-distribution (OOD) variations: a moved goal location, additional target objects, and target objects placed in an unvisited navigation region (see \cref{sec:ood_experiment_details} for details).
For each OOD variation, we evaluate each policy on 20 test configurations.
These settings assess whether the policy can navigate to a relocated goal, handle additional targets, and search beyond demonstrated routes.
\Cref{fig:ood_generalization} shows that \method{SERF} achieves the best task progress across all three variations, suggesting that the explicit spatial representation supports robust behavior under OOD configuration shifts.

\myParagraph{Failure Recovery}
We observe that the \AlgName policy supports recovery from object-drop failures.
We evaluate this capability by comparing \method{SERF} against the image-only \method{PI0.5~(ft)} and the map-conditioned \method{SBP} and \method{SERF~(env)} under the same object-drop procedure.
To induce the failure, we open the gripper during transport, causing the held object to drop and leave the camera view, then resume each policy from the same post-drop state.
Each policy is tested on 20 object-drop episodes.
Since the demonstrations contain no recovery scenarios, successful recovery reflects generalization beyond demonstrated trajectories.
In an example episode, \method{SERF} re-localizes and re-grasps the dropped object, whereas image-only \method{PI0.5~(ft)} fails to recover (\cref{fig:recovery_behavior}, left).
Across all episodes, \method{SERF} outperforms all baselines in both recovery success rate and recovery time, achieving a $95\%$ success rate with an average recovery time of $20.5$\,s (\cref{fig:recovery_behavior}, right).

\newlength{\panelheight}
\setlength{\panelheight}{3.cm} 

\begin{figure}[t]
  \centering

  \begin{minipage}[t]{0.5\linewidth}
    \centering
    \vspace{0pt}

    \begin{minipage}[t][\panelheight][c]{\linewidth}
      \centering
      \small
      \renewcommand{\arraystretch}{1.1}
      \setlength{\tabcolsep}{3pt}

      \begin{tabularx}{\linewidth}{lYYY}
        \toprule
        Method & Task 21 & Task 22 & Task 26 \\
        \midrule
        \method{PI0.5~(pre)}~\cite{larchenko2025task}
          & 42.9{\scriptsize $\pm$19.7} & 43.0{\scriptsize $\pm$23.5} & 44.1{\scriptsize $\pm$22.4} \\
        \method{PI0.5~(ft)}~\cite{larchenko2025task}
          & 40.7{\scriptsize $\pm$18.8} & 43.0{\scriptsize $\pm$19.3} & 48.4{\scriptsize $\pm$21.1} \\
        \method{SBP}~\cite{kim2025seeing}
          & 57.9{\scriptsize $\pm$17.8} & 52.5{\scriptsize $\pm$24.1} & 51.6{\scriptsize $\pm$11.7} \\
        \rowcolor{gray!15}
        \method{SERF~(env)}
          & 57.9{\scriptsize $\pm$15.3} & 59.0{\scriptsize $\pm$20.5} & 49.4{\scriptsize $\pm$13.4} \\
        \rowcolor{gray!15}
        \method{SERF}
          & \textbf{63.5}{\scriptsize $\pm$16.7} & \textbf{60.1}{\scriptsize $\pm$19.1} & \textbf{52.5}{\scriptsize $\pm$13.8} \\
        \bottomrule
      \end{tabularx}
    \end{minipage}

    \captionof{table}{
      Task progress (\%) across BEHAVIOR-1K tasks.
      All methods are built on the same base policy, $\pi_{0.5}$~\cite{black2025pi05}. All fine-tuned methods use the same training setup.
    }
    \label{tab:quantitative_result}
  \end{minipage}
  \hfill
  \begin{minipage}[t]{0.48\linewidth}
    \centering
    \vspace{0pt}

    \begin{minipage}[t][\panelheight][c]{\linewidth}
      \centering
      \includegraphics[width=\linewidth]{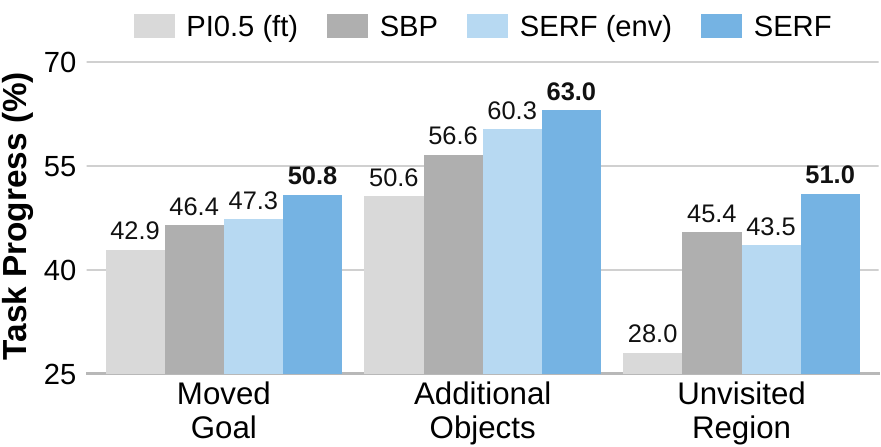}
    \end{minipage}

    \captionof{figure}{
      Scene-configuration generalization under OOD settings.
      We report task progress (\%) for each policy under test-time scene shifts.
    }
    \label{fig:ood_generalization}
  \end{minipage}
\end{figure}

\begin{figure*}
  \centering
  \includegraphics[width=\textwidth]{fig/failure_new.pdf}
  \caption{
    Failure recovery after object drop during transport.
    \textbf{Left:} \method{SERF} re-localizes and re-grasps the dropped object, whereas image-only \method{PI0.5~(ft)} fails to recover.
    \textbf{Right:} \method{SERF} outperforms all baselines in both recovery success rate and recovery time.
    }
    \label{fig:recovery_behavior}
\end{figure*}

\section{Conclusion}
\label{sec:conclusion}

Long-horizon mobile manipulation requires continual reasoning about localization, environment changes, and task progress, all of which are challenging to infer from short observation sequences. We introduced \AlgName, a spatiotemporal feature map that represents the environment and the robot body as neural points in a shared latent space. We showed that conditioning a VLA policy on tokens extracted from the \AlgName map allows the policy to leverage both allocentric scene memory and egocentric robot--environment reasoning. 
As a result, the \AlgName policy outperforms image-only baselines on long-horizon mobile manipulation tasks, improving task progress, trajectory efficiency, robustness to scene-configuration shifts, and recovery from object-drop failures.

\section*{Limitations}
\label{sec:limitations}

\AlgName has several limitations.
First, \AlgName currently relies on a prior map whose features are learned before execution and on privileged instance labels and robot base poses from simulation for map construction and updates.
Accordingly, comparisons with image-only VLA baselines should be interpreted as evaluating the benefit of adding persistent spatial memory under these assumptions, rather than as a controlled comparison between policies with exactly the same input signals.
A feed-forward encoder could initialize neural point features from streaming observations and reduce the dependence on a prior map~\cite{tian2025miso}.
Initial experiments in \cref{sec:privileged_signals} further suggest that these privileged inputs can be replaced at test time by estimates from onboard perception.
Second, real-world mobile manipulation experiments are needed to assess transfer beyond simulation.
Third, our temporal map updates assume object-level rigid motion, limiting updates to scene changes that can be approximated by $\SE(3)$ transforms.
Articulated objects could be handled by extending the formulation to part-level $\SE(3)$ motion using part-level segmentation, while deformable objects require modeling non-rigid dynamics, potentially using learned dynamics models (\eg, \cite{huang2026pointworld}).
Fourth, incorporating map tokens from multiple temporal windows, including past observations and predicted future states, could further strengthen temporal reasoning.
Finally, our current map tokenizer uses manually specified spatial subsets and does not explicitly separate task-level semantics from robot, object, and scene-state cues.
Future task-conditioned tokenization could produce more semantically disentangled map tokens, improving robustness in long-horizon mobile manipulation.

\section*{Acknowledgments}
We gratefully acknowledge support from NSF CCF-2402689 (ExpandAI), NSF 2120019 (CHASE-CI), and the Agency for Defense Development grant funded by the Korean Government (912A45701).


\bibliography{references}

\clearpage
\newgeometry{
  textheight=9.5in,
  textwidth=7in,
  top=0.75in,
  left=0.75in,
  columnsep=0.2in
}
\setlength{\columnsep}{0.2in}
\sloppy
\flushbottom
\appendix
\nolinenumbers
\crefalias{section}{appendix}
\crefalias{subsection}{appendix}
\crefalias{subsubsection}{appendix}
\twocolumn
\section*{Appendix}
\begingroup
\setlength{\parindent}{0pt}
\setlength{\parskip}{0.25em}

\newcommand{\appendixtocsection}[2]{%
  \hyperref[#1]{\textbf{\ref*{#1}\quad #2}}\dotfill\pageref{#1}\par
}

\appendixtocsection{sec:privileged_signals}{Privileged Inputs and Onboard Estimates}
\vspace{0.35em}

\appendixtocsection{sec:coordinate_feature_pairs}{Map Dataset Generation}
\vspace{0.35em}

\appendixtocsection{sec:map_representation_details}{Map Representation Details}
\vspace{0.35em}

\appendixtocsection{sec:contrastive_losses}{Contrastive Objectives}
\vspace{0.35em}

\appendixtocsection{sec:map_update_details}{Map Updates and Tracking}
\vspace{0.35em}

\appendixtocsection{sec:vla_policy_details}{VLA Policy Details}
\vspace{0.35em}

\appendixtocsection{sec:map_token_ablation}{Map Token Ablation}
\vspace{0.35em}

\appendixtocsection{sec:behavior_1k_details}{BEHAVIOR-1K Benchmark Details}
\vspace{0.35em}


\appendixtocsection{sec:image_history}{Comparison with Image-History Baseline}
\vspace{0.35em}

\appendixtocsection{sec:time_to_subgoal}{Task Progress Curves}
\vspace{0.35em}

\appendixtocsection{sec:ood_experiment_details}{Scene-Configuration Experiments}
\vspace{0.35em}

\appendixtocsection{sec:more_qualitative_results}{Additional Qualitative Results}
\vspace{0.35em}

\appendixtocsection{sec:more_map_visualizations}{Additional Map Visualizations}

\endgroup

\section{Privileged Inputs and Onboard Estimates}
\label{sec:privileged_signals}

\begin{figure}[!htbp]
  \centering
  \includegraphics[width=\linewidth]{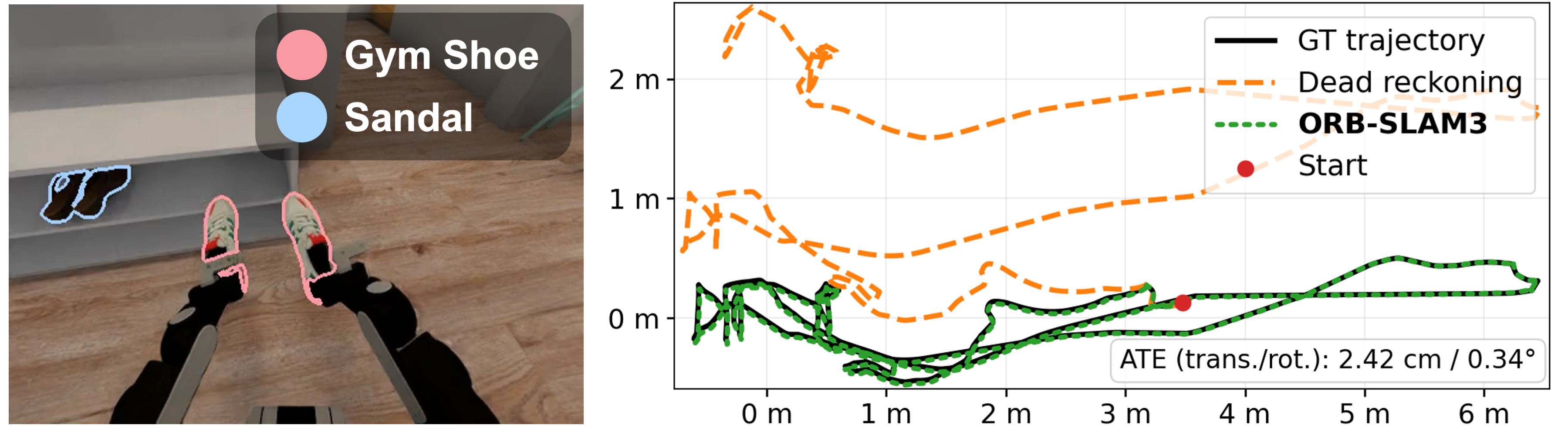}
  \caption{
    \textbf{Left:} Text-prompted SAM~3 masks for the target object categories on Task~22.
    \textbf{Right:} ORB-SLAM3 base pose estimates compared with the ground truth and dead reckoning.
  }
  \label{fig:appendix_privileged}
\end{figure}

\AlgName uses instance segmentation labels and robot base poses as privileged inputs that the image-only baselines do not use.
To assess feasibility with onboard perception, we replace both inputs at test time.
Instance labels are derived from text-prompted SAM~3 masks~\cite{carion2026sam3} (\cref{fig:appendix_privileged}, left), where the prompts are the target object categories in the BDDL specification (\eg, \emph{gym shoe} and \emph{sandal}).
Base poses are estimated by ORB-SLAM3~\cite{campos2021orbslam3} from the head-camera observations (\cref{fig:appendix_privileged}, right).
On Task~22, ORB-SLAM3 estimates closely follow the ground-truth trajectory, with an absolute trajectory error of $2.42\,\mathrm{cm}$ in translation and $0.34^\circ$ in rotation, whereas dead reckoning drifts substantially.
\Cref{tab:appendix_privileged} shows that estimated inputs reduce task progress by $5.3$ points relative to ground-truth inputs.
Even with estimated inputs, \AlgName achieves $11.8$ points higher task progress than the image-only \method{PI0.5~(ft)} baseline.

\begin{table}[H]
    \centering
    \small
    \renewcommand{\arraystretch}{1.15}
    \begin{tabularx}{\linewidth}{lY}
        \toprule
        Method & Task 22 \\
        \midrule
        \method{PI0.5~(ft)}~\cite{larchenko2025task}
            & 43.0{\scriptsize $\pm$19.3} \\
        \rowcolor{gray!15}
        \method{SERF}~(estimated labels/poses)
            & 54.8{\scriptsize $\pm$22.9} \\
        \rowcolor{gray!15}
        \method{SERF}~(GT labels/poses)
            & \textbf{60.1}{\scriptsize $\pm$19.1} \\
        \bottomrule
    \end{tabularx}
    \vspace{0.4em}
    \caption{
        \textbf{Task progress (\%) on Task~22.}
        \method{SERF} is evaluated with estimated or ground-truth (GT) instance labels and base poses.
    }
    \label{tab:appendix_privileged}
\end{table}

\section{Map Dataset Generation}
\label{sec:coordinate_feature_pairs}

\myParagraph{Environment Samples}
We construct the reconstruction dataset for \cref{prob:map} as coordinate--embedding pairs $(x,y)\in\calX\times\calY$.
We are given an RGB image $I$, a depth image $Z$, and the camera-to-world pose $(R_{\text{cam}}, t_{\text{cam}})$. 
The VFM encoder produces per-patch embeddings $Y \in \calY^{H \times W}$ from $I$.
We back-project each patch center $\rho$ into the world frame:
\begin{equation}
\label{eq:backproj}
    x(\rho) = R_{\text{cam}}\,\Pi^{-1}
    \begin{bmatrix} \rho \\ 1 \end{bmatrix}
    Z[\rho] \, + \, t_{\text{cam}}.
\end{equation}
Here, $\Pi^{-1}$ back-projects using camera intrinsics, and $(R_{\text{cam}}, t_{\text{cam}})$ maps the point from the camera frame to the world frame.
Each patch yields a pair $(x(\rho), y(\rho))$, where $y(\rho)=Y[\rho]$.
These pairs form the environment samples.

\begin{figure}[!htbp]
  \centering
  \includegraphics[width=\linewidth]{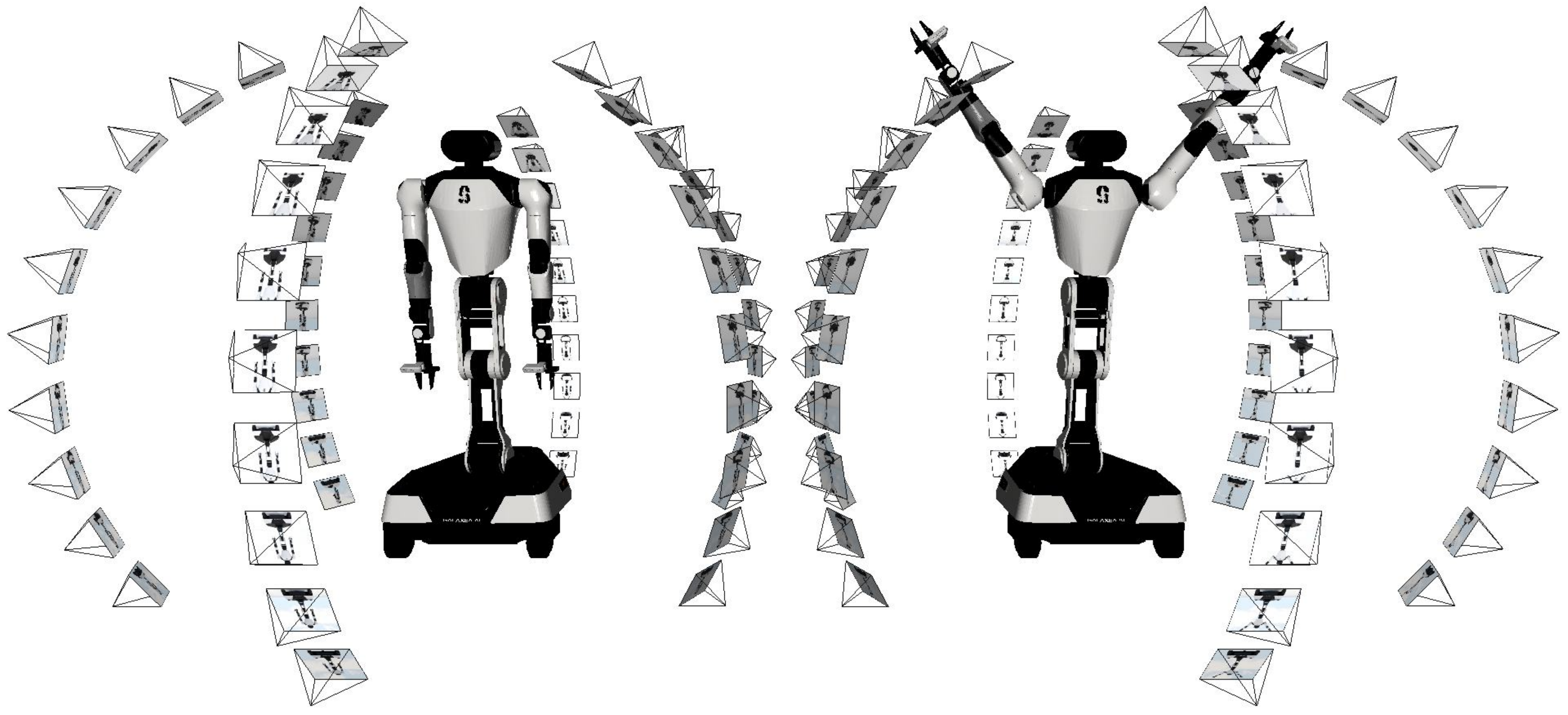}
  \caption{
    Robot renderings from multiple viewpoints provide state-conditioned reconstruction samples.
  }
  \label{fig:robot_mapping_dataset}
\end{figure}

\myParagraph{Robot Samples}
Reconstructing the robot body also uses coordinate--embedding pairs.
Robot samples are generated from a fixed set of robot states selected to cover representative base, arm, and gripper configurations (see \cref{fig:robot_mapping_dataset}).
For each selected state $s$, we render the robot from multiple viewpoints, extract patch embeddings $Y_s$, and back-project robot-body patches using \eqref{eq:backproj} to obtain $x_s(\rho)\in\calX$; see \cref{fig:robot_mapping_dataset}.
Each patch contributes a pair $(x_s(\rho), y_s(\rho))$, where $y_s(\rho)=Y_s[\rho]$.
Aggregating these pairs across viewpoints and states yields the robot samples.

\section{Map Representation Details}
\label{sec:map_representation_details}

We use DINOv3~\cite{simeoni2025dinov3} to obtain per-patch VFM embeddings.
For the $480 \times 480$ RGB observations, DINOv3 produces $1280$-dimensional embeddings on a $30 \times 30$ grid with a patch size of $16$.
We obtain part-level masks using the SAM 2 base-plus model~\cite{ravi2024sam}.
We register environment neural points at a voxel size of $0.02\,\mathrm{m}$ and sample robot neural points from the robot mesh surface at the same resolution.
Each neural point stores a $64$-dimensional latent feature.
The shared decoder $D_\theta$ is a residual MLP mapping $64$-dimensional neural point features to $1280$-dimensional DINOv3 patch embeddings.
For each spatial query, we collect candidate neural points via a ball query, select the $K{=}6$ nearest neighbors, and interpolate their latent features using a softmax temperature of $0.05$, as in \eqref{eq:latent_feature_query}.
\section{Contrastive Objectives}
\label{sec:contrastive_losses}

\begin{figure}[!htbp]
  \centering
  \includegraphics[width=\linewidth]{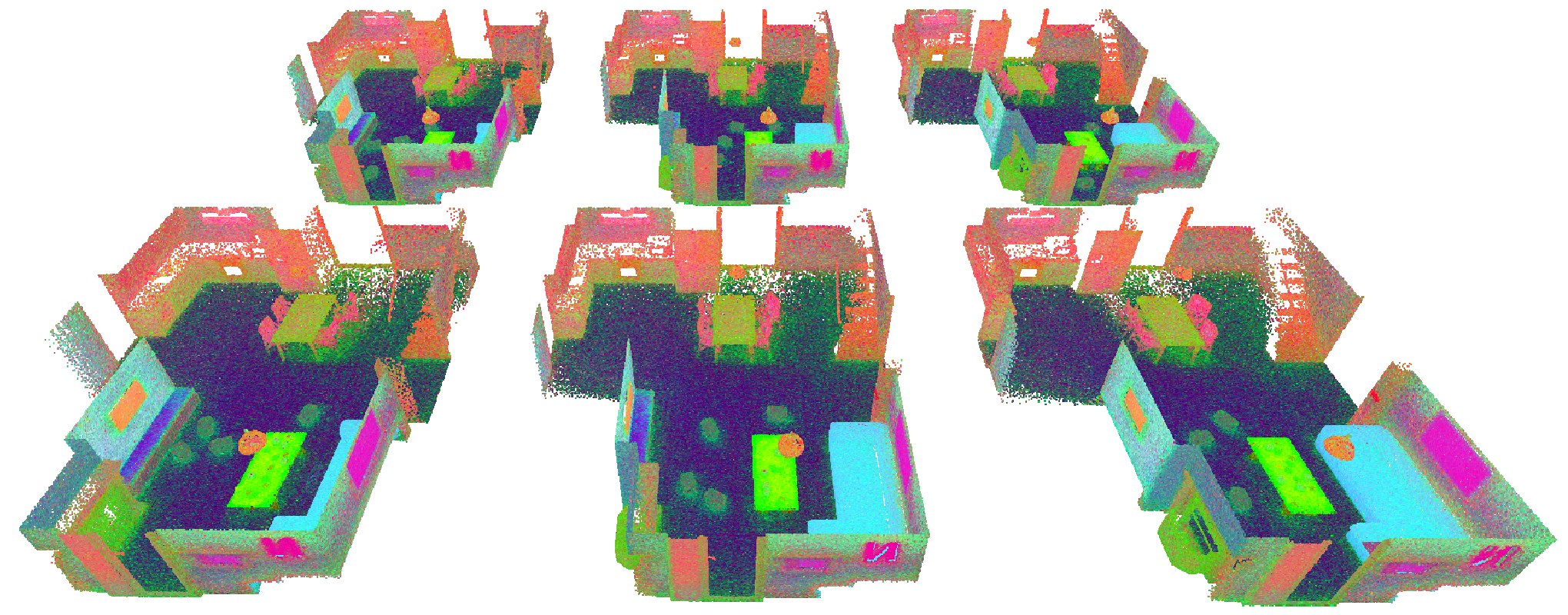}
  \caption{
    PCA projections show same-category features clustering across scenes and different-category features separating.
  }
  \label{fig:inter_contrastive}
\end{figure}

\myParagraph{Inter-Category Objective}
The inter-category contrastive objective pulls same-category features together and separates different-category features~\cite{oord2018representation}:
\begin{equation}
\label{eq:inter}
    \calL_{\text{inter}} = -\log \frac{\exp \bigl(\mathrm{sim}(f_i, f_j^{+})/\sigma_{\mathrm{c}}\bigr)}{\sum_{k} \exp \bigl(\mathrm{sim}(f_i, f_k)/\sigma_{\mathrm{c}}\bigr)}.
\end{equation}
Here, $f_j^{+}$ is a same-category positive for $f_i$, and the denominator includes positives and negatives from all categories.
For cross-scene training, we construct a category-indexed feature bank from the neural points and sample up to $16{,}384$ points per iteration.
Category-balanced importance sampling draws an equal number of points from each category for the contrastive batch.
We treat robot features as an additional category.
We set $\lambda_{\mathrm{inter}} = 0.02$ and $\sigma_{\mathrm{c}} = 0.1$.
\Cref{fig:inter_contrastive} visualizes the category-level structure across multiple scenes using PCA.

\begin{figure}[!htbp]
  \centering
  \includegraphics[width=\linewidth]{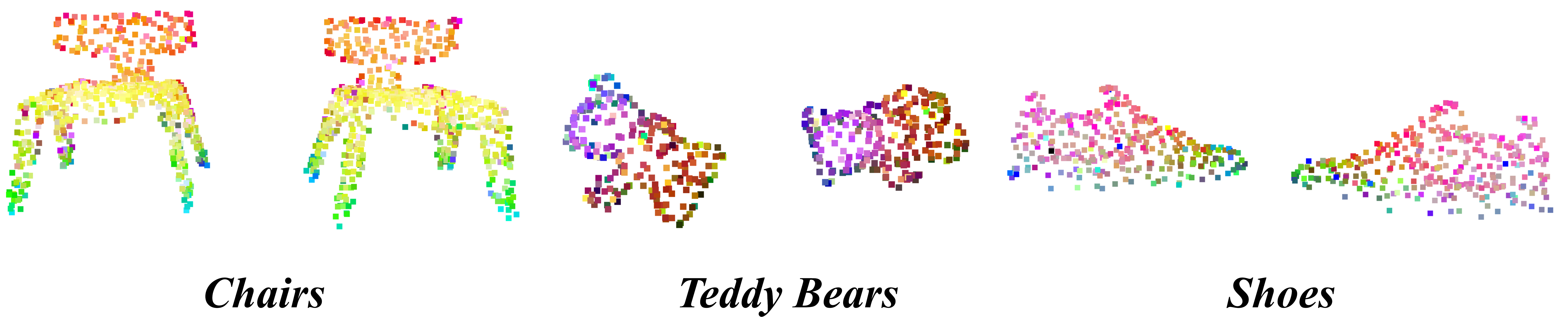}
  \caption{
    PCA projections show same-part features clustering together and different-part features separating within each instance.
  }
  \label{fig:intra_contrastive}
\end{figure}

\myParagraph{Intra-Instance Objective}
The intra-instance contrastive objective pulls same-part features together and separates different-part features~\cite{oord2018representation}:
\begin{equation}
\label{eq:intra}
    \calL_{\text{intra}} = -\log \frac{\exp \bigl(\mathrm{sim}(f_i, f_j^{+})/\sigma_{\mathrm{c}}\bigr)}{\sum_{k \in \text{parts}} \exp \bigl(\mathrm{sim}(f_i, f_k)/\sigma_{\mathrm{c}}\bigr)}.
\end{equation}
Here, $(f_i, f_j^{+})$ is a same-part positive pair, and the denominator includes sampled features from all part labels within the same instance.
This encourages the latent features to capture part-level structure useful for manipulation.
Environment samples use SAM 2~\cite{ravi2024sam} segments as part labels, while robot samples use labels rendered from robot links.
We sample up to $16{,}384$ points per iteration with category-balanced importance sampling and set $\lambda_{\mathrm{intra}} = 0.01$ and $\sigma_{\mathrm{c}} = 0.1$.
\Cref{fig:intra_contrastive} visualizes the learned part-level structure using PCA.

\section{Map Updates and Tracking}
\label{sec:map_update_details}

\myParagraph{Map Updates}
Policy training and online execution use the same map update rule: object-level $\SE(3)$ transforms update environment points, while forward kinematics updates robot points.
The two settings differ only in whether the robot states and object-level transforms used for these updates are available offline or estimated online.
During training, recorded robot states and precomputed object-level $\SE(3)$ trajectories from demonstrations are used to position robot points and move environment points, respectively.
During execution, robot points are positioned using the online robot state, and object-level $\SE(3)$ transforms are estimated from robot observations before each policy step.

\myParagraph{Object Tracking}
We track only movable object instances and do not estimate $\SE(3)$ transforms for stationary objects (\eg, tables).
To estimate object-level $\SE(3)$ transforms, we track 2D keypoints within each instance mask and lift them to 3D using depth and camera pose.
For each instance, we initialize 2D keypoints at Shi-Tomasi corners~\cite{shi1994good} within the instance mask and track them with CoTracker3~\cite{karaev2024cotracker}.
We maintain a separate CoTracker buffer for each head and wrist camera view.
If the 3D displacement of the instance centroid between tracking steps is less than $0.015\,\mathrm{m}$, we treat the instance as stationary and skip registration for that step.
To recover from tracking failures (\eg, after occlusion or rapid motion), if the observed instance centroid deviates from the stored centroid by more than $0.20\,\mathrm{m}$, we bypass FGR and initialize ICP directly from the centroid offset.
We accept this update only if the resulting centroid error is at most $0.02\,\mathrm{m}$.
\section{VLA Policy Details}
\label{sec:vla_policy_details}

\vspace{-5pt}
\myParagraph{Map Tokenizer}
At each policy step, we filter the map to retain robot points and task-relevant object points specified by the BDDL task, while removing background and structural points.
For each policy input, we sample $25{,}000$ neural points from the filtered point set.
The tokenizer consists of a shared two-stage Point Transformer backbone, eight branch-specific Point Transformer heads, and per-branch attention pooling.
The shared backbone has channel widths $(128, 256)$, two Point Transformer blocks per stage, stride $4$ at each stage, and $16$ nearest neighbors for local attention.
Each branch-specific head has width $256$, stride $1$, and $16$ nearest neighbors.
Per-branch attention pooling maps each branch output to one $2048$-dimensional map token, matching the VLA token dimension.

\myParagraph{VLA Training and Inference}
We use the BEHAVIOR-1K OpenPI implementation of Larchenko \etal~\cite{larchenko2025task} as the base VLA model.
We use an action horizon of $30$ and an action dimension of $32$.
For each task, we fine-tune the model for $20k$ steps with a batch size of $16$, using $15$ flow-matching time/noise samples per batch element.
The learning rate follows a cosine schedule with $1k$ warmup steps, a peak learning rate of $2.5{\times}10^{-6}$, and a final learning rate of $1.0{\times}10^{-6}$.
During evaluation, each policy query produces a $30$-step action chunk, so we recompute map tokens once per query rather than at every low-level control step.
The policy generates each action chunk with $20$ Euler integration steps.
Following the execution protocol, we keep the first $26$ actions from each $30$-step chunk, apply cubic interpolation, and resample them into $20$ control commands.

\section{Map Token Ablation}
\label{sec:map_token_ablation}

\vspace{-5pt}
\Cref{tab:map_token_ablation} reports an ablation over the five map-token groups used by the map tokenizer.
For each variant, we omit one map-token group while keeping the policy model and training setup fixed.
We evaluate all variants on Task~22. 
The full \method{SERF} model achieves the highest task progress.

\begin{table}[t]
    \centering
    \small
    \renewcommand{\arraystretch}{1.1} 
    \begin{tabularx}{\linewidth}{lY}
        \toprule
        Method & Task 22 \\
        \midrule
        \method{SERF w/o robot-base}
            & 56.5{\scriptsize $\pm$21.7} \\
        \method{SERF w/o end-effector}
            & 54.0{\scriptsize $\pm$21.5} \\
        \method{SERF w/o robot-only}
            & 54.0{\scriptsize $\pm$22.2} \\
        \method{SERF w/o environment-only}
            & 58.5{\scriptsize $\pm$20.9} \\
        \method{SERF w/o global}
            & 55.1{\scriptsize $\pm$20.6} \\
        \rowcolor{gray!15}
        \method{SERF}
            & \textbf{60.1}{\scriptsize $\pm$19.1} \\
        \bottomrule
    \end{tabularx}
    \vspace{0.4em}
    \caption{
        \textbf{Map token ablation.}
        Each variant omits one map-token group. We report task progress (\%).
    }
    \label{tab:map_token_ablation}
    \vspace{-20pt}
\end{table}

\section{BEHAVIOR-1K Benchmark Details}
\label{sec:behavior_1k_details}

\vspace{-5pt}
BEHAVIOR-1K~\cite{li2022behavior} is a benchmark of long-horizon household bimanual mobile manipulation tasks in OmniGibson, with goals specified in BDDL.
We evaluate three tasks from the benchmark.
For each task, we train on $200$ expert demonstrations and evaluate on $20$ configurations.

\begin{table}[!htbp]
\vspace{-5pt}
\centering
\small
\setlength{\tabcolsep}{4pt}
\renewcommand{\arraystretch}{1.}
\begin{tabularx}{\linewidth}{l X c c}
\toprule
Task & Target objects & Workspace ($x{\times} y$) & Max steps \\
\midrule
Task 21 & 2 dice, 2 teddy bears, 2 board games, 1 toy train & $4.4{\times}3.0$\,m & 38,372 \\
Task 22 & 2 gym shoes, 2 sandals & $5.5{\times}9.0$\,m & 15,384 \\
Task 26 & 4 baskets, 4 candles, 4 butter cookies, 4 Swiss cheeses, 4 bows & $7.6{\times}8.8$\,m & 52,120 \\
\bottomrule
\end{tabularx}
\vspace{0.4em}
\caption{\textbf{BEHAVIOR-1K task details.}}
\label{tab:behavior_1k_task_details}
\vspace{-15pt}
\end{table}

\myParagraph{Task Summary}
\Cref{tab:behavior_1k_task_details} summarizes each task's target objects, task-relevant workspace extent, and maximum rollout length.
Workspace extent denotes the $x$- and $y$-dimensions of the task-relevant region.
We set the maximum rollout length in evaluation to twice the average demonstration length.

\myParagraph{BDDL Goal Conditions}
In Task 21, all target toys must be placed in the same bookcase.
In Task 22, each footwear item must be placed on the rack without touching the floor, and items of the same category must be next to each other.
In Task 26, each basket must contain one candle, one butter cookie, one Swiss cheese, and one bow.

\section{Comparison with Image-History Baseline}
\label{sec:image_history}
\vspace{-5pt}

Conditioning on observation history is another way to provide memory for long-horizon tasks~\cite{sridhar2025memer,torne2026mem,koo2026hamlet}.
\Cref{tab:image_history_comparison} compares \AlgName with an image-history baseline, \method{PI0.5~(image history)}.
The baseline uses the same $\pi_{0.5}$ policy and maintains a sliding-window buffer of eight past frames sampled at a fixed stride.
A memory module~\cite{koo2026hamlet} encodes these frames into eight history tokens for policy conditioning.
Although \method{PI0.5~(image history)} improves task progress over \method{PI0.5~(ft)} on Tasks~21 and~22, \AlgName outperforms it on all three tasks.

\begin{table}[!htbp]
  \centering
  \small
  \renewcommand{\arraystretch}{1.1}
  \setlength{\tabcolsep}{5pt}

  \begin{tabularx}{\linewidth}{lYYY}
    \toprule
    Method & Task 21 & Task 22 & Task 26 \\
    \midrule
    \method{PI0.5~(ft)}~\cite{larchenko2025task}
      & 40.7{\scriptsize $\pm$18.8}
      & 43.0{\scriptsize $\pm$19.3}
      & 48.4{\scriptsize $\pm$21.1} \\
    \method{PI0.5~(image history)}~\cite{koo2026hamlet}
      & 50.0{\scriptsize $\pm$17.2}
      & 49.5{\scriptsize $\pm$19.4}
      & 48.1{\scriptsize $\pm$20.2} \\
    \method{SBP}~\cite{kim2025seeing}
      & 57.9{\scriptsize $\pm$17.8}
      & 52.5{\scriptsize $\pm$24.1}
      & 51.6{\scriptsize $\pm$11.7} \\
    \cellcolor{gray!15}\method{SERF}
      & \cellcolor{gray!15}\textbf{63.5}{\scriptsize $\pm$16.7}
      & \cellcolor{gray!15}\textbf{60.1}{\scriptsize $\pm$19.1}
      & \cellcolor{gray!15}\textbf{52.5}{\scriptsize $\pm$13.8} \\
    \bottomrule
  \end{tabularx}
  \vspace{0.4em}

  \caption{
    \textbf{Comparison with an image-history baseline.}
    We report task progress (\%).
  }
  \label{tab:image_history_comparison}
  \vspace{-20pt}
\end{table}

\section{Task Progress Curves}
\label{sec:time_to_subgoal}

\begin{figure}[!htbp] 
    \centering
    \includegraphics[width=0.495\linewidth]{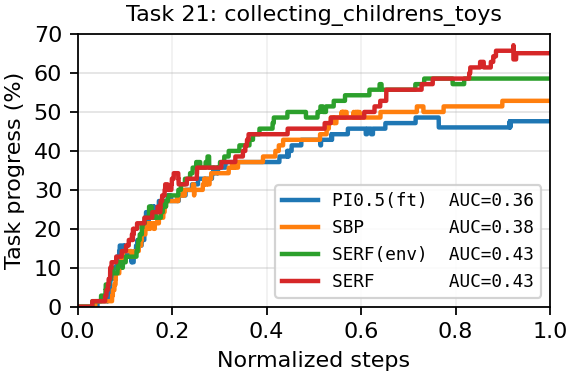}
    \hfill
    \includegraphics[width=0.495\linewidth]{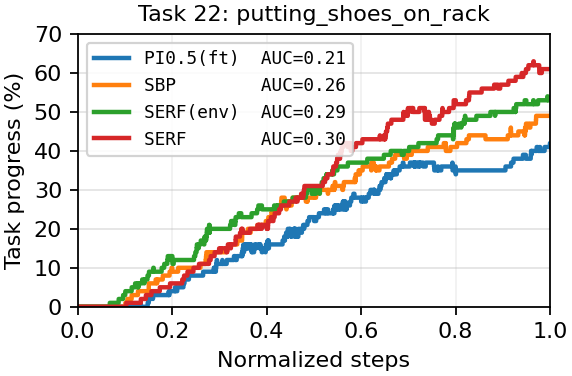}
    \caption{
        Task progress over execution steps on Tasks~21 and~22. \method{SERF} achieves the highest AUC and final task progress on both tasks.
    }
    \label{fig:qscore_curves}
\end{figure}

We report per-step task progress curves and their AUCs,
averaged over 10 configurations, for Tasks~21 and~22 (see \cref{fig:qscore_curves}).
Each curve shows how task progress evolves over execution steps, while the AUC summarizes the progress accumulated over the episode.
On both tasks, \AlgName achieves the highest AUC, indicating faster progress, and reaches the highest final task progress.

\section{Scene-Configuration Experiments}
\label{sec:ood_experiment_details}

In all three out-of-distribution (OOD) experiments, policies are trained only on the original in-distribution scenes.
OOD variations appear only during evaluation and test generalization to changes in the goal position, the number of target objects, and the placement of objects in previously unvisited regions.
We evaluate each variation on $20$ configurations.

\begin{figure}[!htbp]
  \centering
  \includegraphics[width=\linewidth]{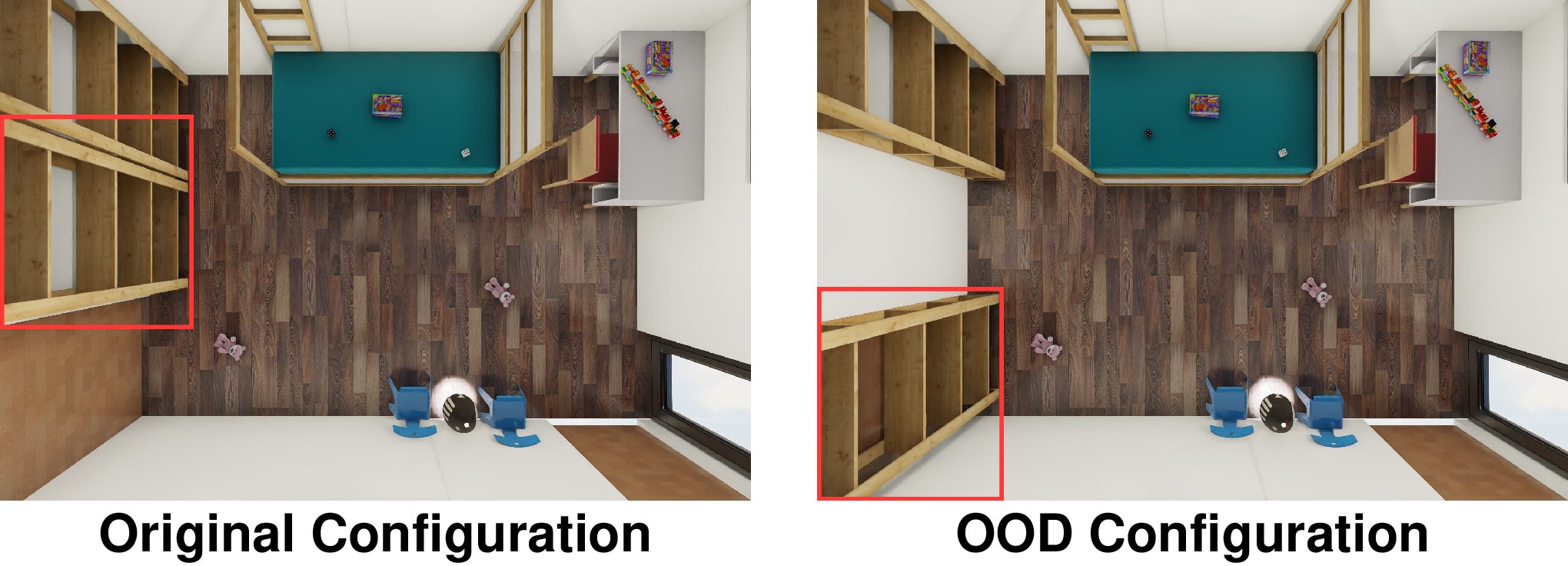}
  \caption{
     The goal bookcase is relocated from its original position.
  }
  \label{fig:appendix_goal_ood}
  \vspace{-5pt}
\end{figure}

\myParagraph{Moved Goal}
We evaluate the \emph{Moved Goal} variation on Task~21, where the bookcase serves as the placement goal for the collected toys (see \cref{fig:appendix_goal_ood}).
In this variation, we relocate the bookcase while keeping the target objects, language instruction, and BDDL goal conditions unchanged.
This tests whether the policy uses the updated scene map to navigate to the new goal location instead of relying on memorized demonstration locations.

\begin{figure}[!htbp]
  \centering
  \includegraphics[width=\linewidth]{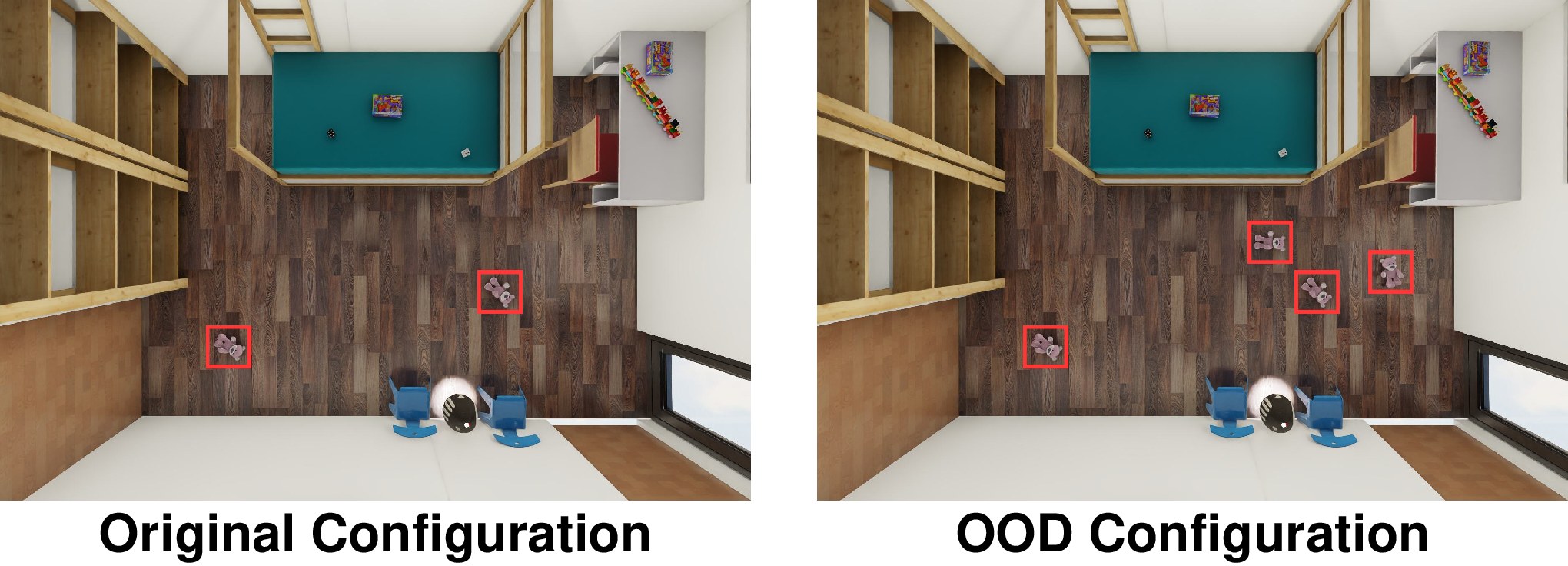}
  \caption{
    Additional teddy bears are added as target objects in the original scene.
  }
  \label{fig:appendix_task_ood}
  \vspace{-5pt}
\end{figure}

\myParagraph{Additional Objects}
We evaluate the \emph{Additional Objects} variation on Task~21.
The original task requires the robot to collect seven target toys and place them on the bookcase.
Adding two teddy bears to the evaluation scene increases the teddy bear targets from two to four, yielding nine target toys in total (see \cref{fig:appendix_task_ood}).
Although the language instruction remains unchanged, the evaluation BDDL goal is updated to include the added teddy bear instances.
This tests whether the policy can identify and transport more target objects than were present in the training demonstrations.

\begin{figure}[!htbp]
  \centering
  \includegraphics[width=\linewidth]{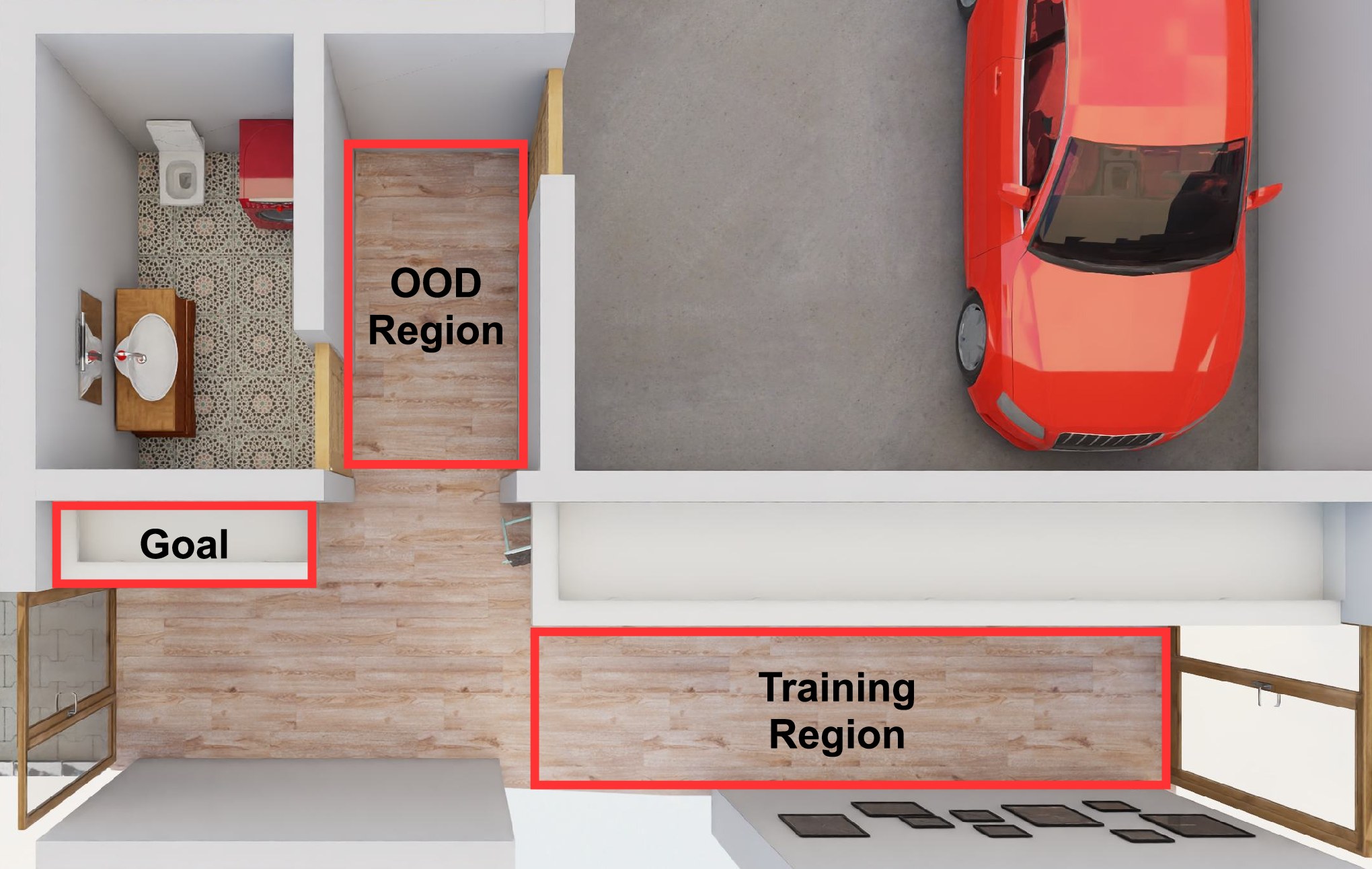}
  \caption{
    During evaluation, two sandals are placed in a region where no target objects appear in the training demonstrations.
  }
  \label{fig:appendix_region_ood}
  \vspace{-10pt}
\end{figure}

\myParagraph{Unvisited Region}
We evaluate the \emph{Unvisited Region} variation on Task~22.
Of the four target items, two sandals are placed in an unvisited region of the same scene (see \cref{fig:appendix_region_ood}).
The language instruction and BDDL goal conditions remain unchanged.
This tests whether the policy can reach an unvisited region by navigating beyond demonstrated routes.

\section{Additional Qualitative Results}
\label{sec:more_qualitative_results}

\begin{figure*}[t]
  \centering
  \includegraphics[width=\textwidth]{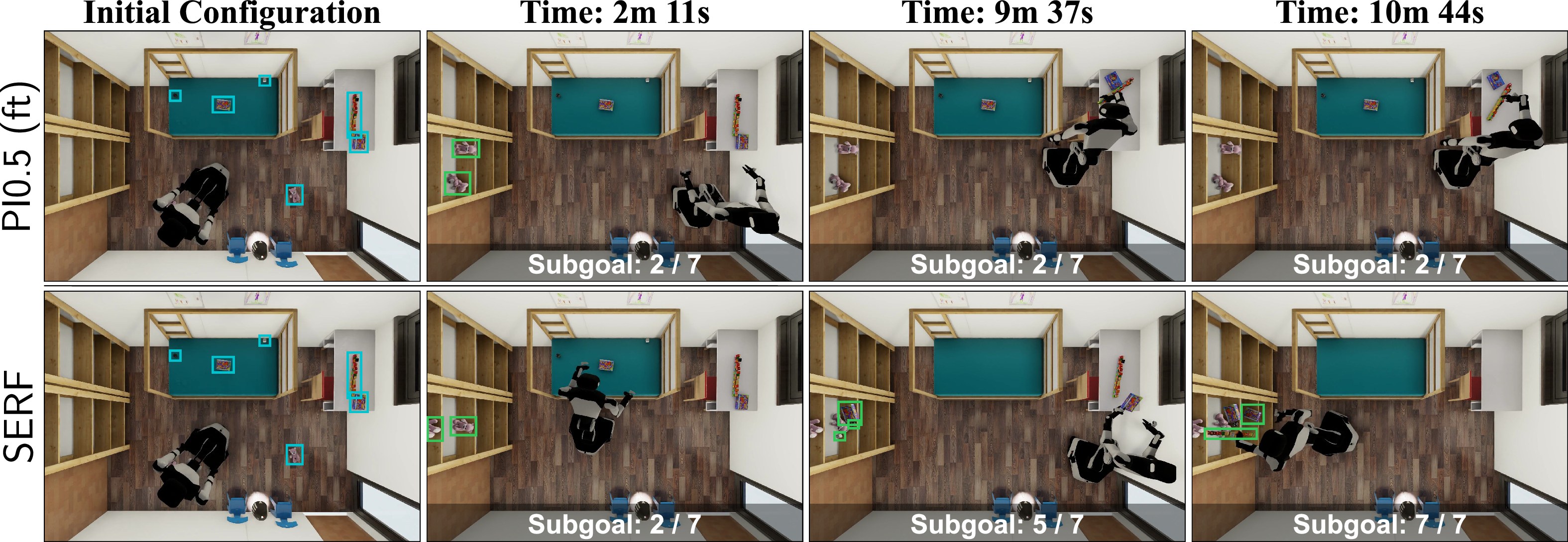}
  \\[0.8em]
  \includegraphics[width=\textwidth]{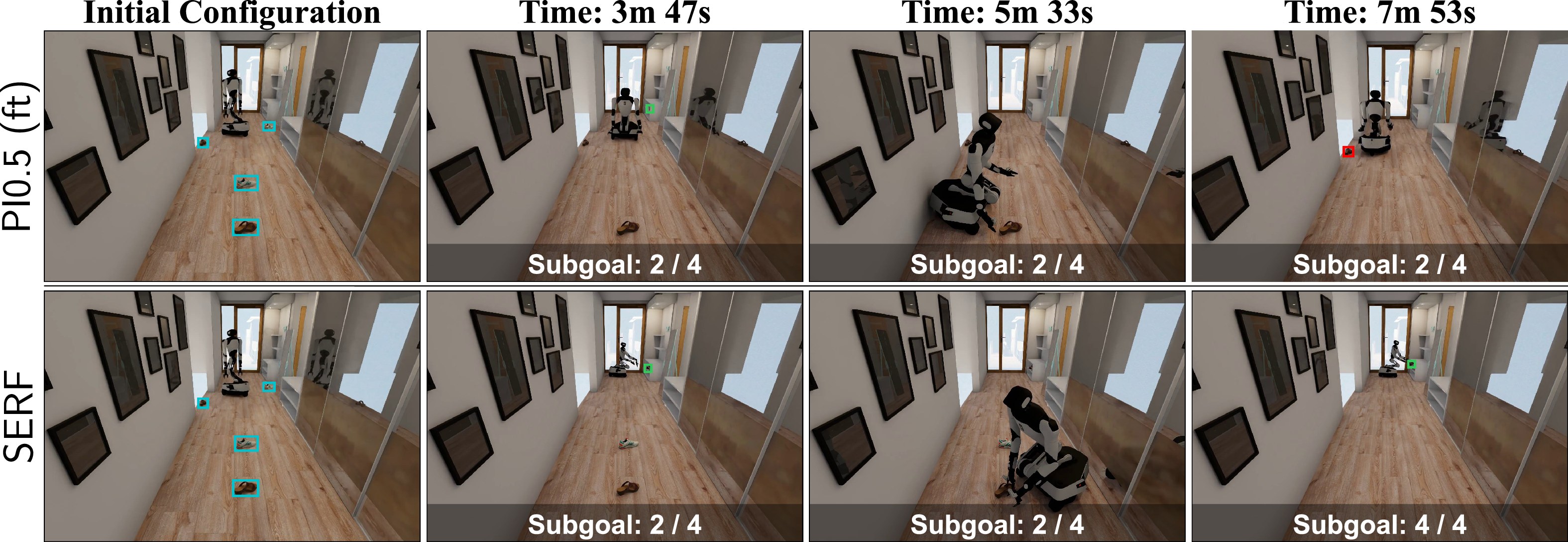}
  \\[0.8em]
  \includegraphics[width=\textwidth]{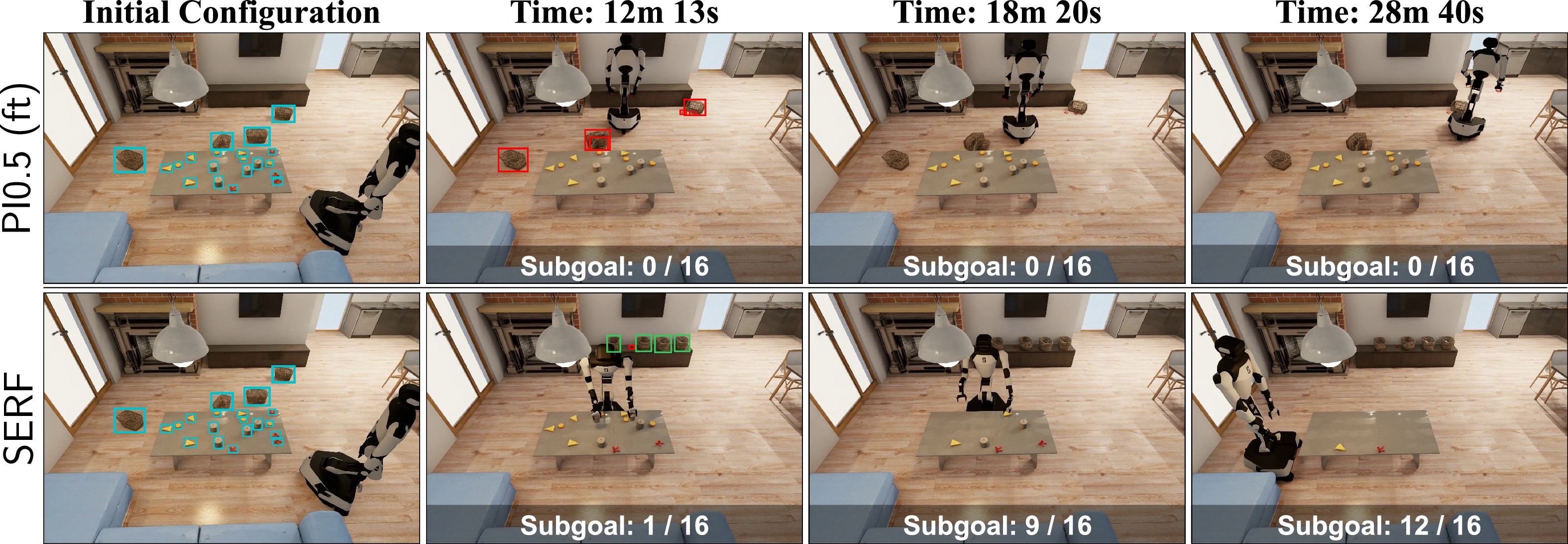}
  \caption{
    \textbf{Qualitative comparisons on long-horizon mobile manipulation.}
    The row pairs correspond to Task~21, Task~22, and Task~26, respectively, and compare \method{PI0.5~(ft)} with \method{SERF}.
    In these rollouts, \method{SERF} achieves higher task progress than \method{PI0.5~(ft)} across all three tasks.
  } 
  \label{fig:additional_qualitative_results}
\end{figure*}

\Cref{fig:additional_qualitative_results} shows qualitative comparisons between \method{PI0.5~(ft)} and \method{SERF} on Tasks~21, 22, and 26.
Across tasks, \method{PI0.5~(ft)} often stalls after task-relevant objects leave the current field of view.
In contrast, \method{SERF} uses the feature map as persistent spatial memory and makes more consistent task progress.
These qualitative results further suggest that map-conditioned policy learning improves long-horizon mobile manipulation.

\section{Additional Map Visualizations}
\label{sec:more_map_visualizations}

\begin{figure*}[t]
  \centering
  \includegraphics[width=\textwidth]{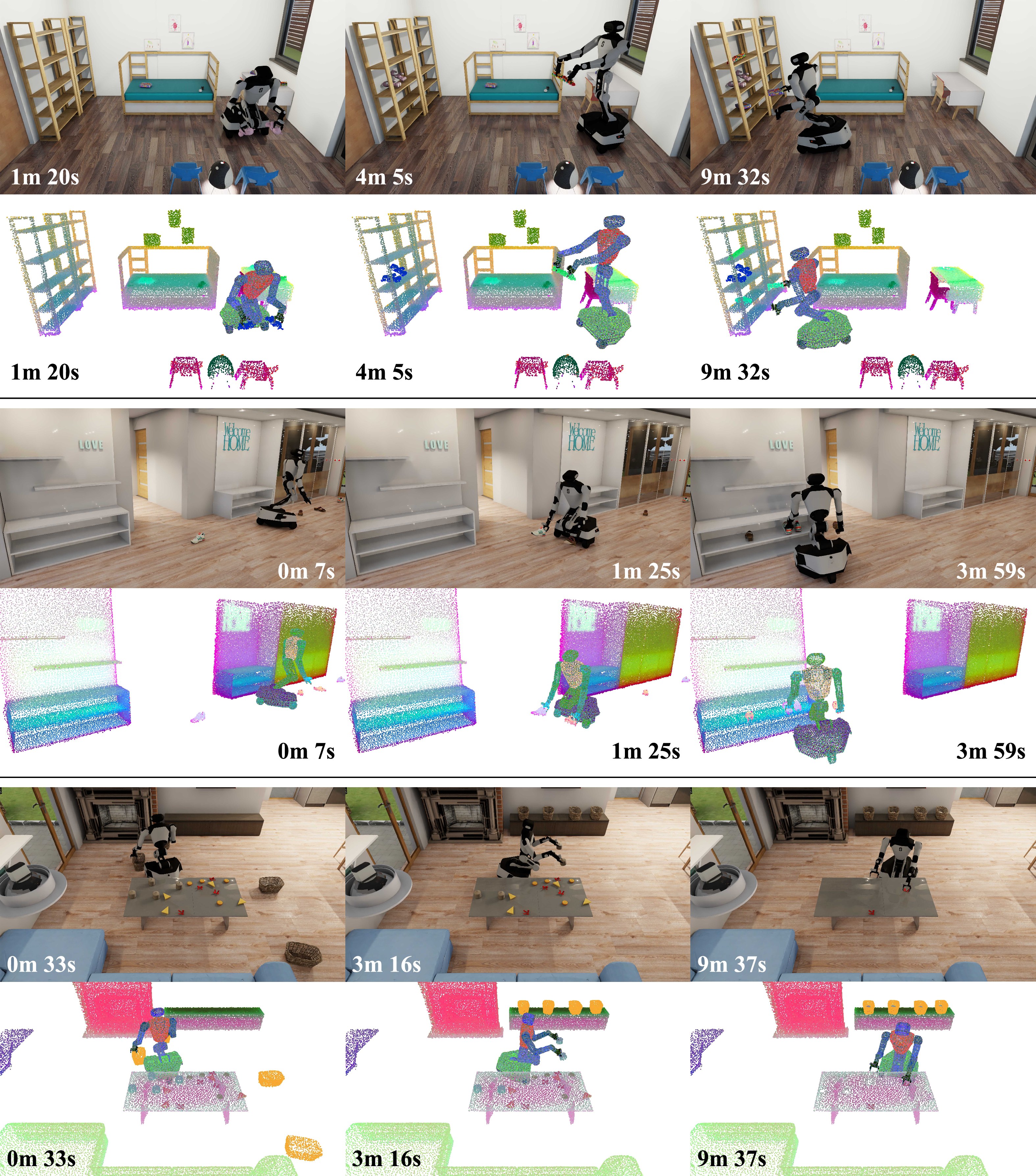}
  \caption{
    \textbf{\AlgName map visualizations.}
    Row pairs correspond to Task~21, Task~22, and Task~26.
    For each task, the top row shows third-person observations of the robot during execution, and the bottom row shows the corresponding \AlgName feature map visualized with PCA.
  }
  \label{fig:additional_map_visualizations}
\end{figure*}

\Cref{fig:additional_map_visualizations} shows \AlgName map visualizations for Tasks~21, 22, and 26.
For each task, the top row shows third-person observations of the robot executing the task, and the bottom row shows the corresponding \AlgName feature map.

\end{document}